\documentclass[journal]{IEEEtran}
\usepackage{ifpdf}
\usepackage{amsthm}
\usepackage{amsmath} 
\newtheorem{theorem}{Theorem}
\theoremstyle{definition}

\newtheorem{remark}{Remark}

\usepackage{xcolor}
\usepackage{hyperref}
\usepackage{graphics} 
\usepackage{epsfig} 
\usepackage{times} 
\usepackage{amsmath} 
\usepackage{amssymb}  

\usepackage{amscd}

\usepackage{graphicx}
\usepackage[ruled,linesnumbered]{algorithm2e}
\usepackage{algpseudocode}
\usepackage{graphics}
\usepackage{booktabs}

\DeclareMathOperator*{\argmax}{argmax}

\usepackage[alpha]{mdpn}  

\usepackage{cite}

\usepackage{graphics} 
\usepackage{epsfig} 
\usepackage{subfigure}
\ifCLASSINFOpdf
\else
  \usepackage[dvips]{graphicx}
\fi
\usepackage{amsmath}
\usepackage{bm}
\usepackage{amssymb}

\usepackage{array}

\usepackage{stfloats}
\ifCLASSOPTIONcaptionsoff
  \usepackage[nomarkers]{endfloat}
 \let\MYoriglatexcaption\caption
    \renewcommand{\caption}[2][\relax]{\MYoriglatexcaption[#2]{#2}}
\fi
\usepackage{url}

\usepackage{fancyhdr}

\begin{document}
%
\title{Collaborative Target Search with a Visual Drone Swarm: An Adaptive Curriculum Embedded Multistage Reinforcement Learning Approach}
%
%
%
    
\author{Jiaping Xiao,~\IEEEmembership{Graduate Student Member,~IEEE},~Phumrapee Pisutsin~and~Mir Feroskhan,~\IEEEmembership{Member,~IEEE}
\thanks{J. Xiao, P. Pisutsin and M. Feroskhan are with the School of Mechanical and Aerospace Engineering, Nanyang Technological University, Singapore 639798, Singapore (e-mail: jiaping001@e.ntu.edu.sg;pisu0001@e.ntu.edu.sg; mir.feroskhan@ntu.edu.sg).
}
}

\markboth{}%
{Shell \MakeLowercase{\textit{et al.}}: Bare Demo of IEEEtran.cls for IEEE Journals}
%



\maketitle

{\color{black}
\begin{abstract}
Equipping drones with target search capabilities is highly desirable for applications in disaster rescue and smart warehouse delivery systems. Multiple intelligent drones that can collaborate with each other and maneuver among obstacles show more effectiveness in accomplishing tasks in a shorter amount of time. However, carrying out collaborative target search (CTS) without prior target information is extremely challenging, especially with a visual drone swarm. In this work, we propose a novel data-efficient deep reinforcement learning approach called Adaptive Curriculum Embedded Multistage Learning (ACEMSL) to address these challenges, mainly 3D sparse reward space exploration with limited visual perception and collaborative behavior requirements. Specifically, we decompose the CTS task into several subtasks including individual obstacle avoidance, target search, and inter-agent collaboration, and progressively train the agents with multistage learning. Meanwhile, an adaptive embedded curriculum is designed, where the task difficulty level can be adaptively adjusted based on the success rate achieved in training. ACEMSL allows data-efficient training and individual-team reward allocation for the visual drone swarm. Furthermore, we deploy the trained model over a real visual drone swarm and perform CTS operations without fine-tuning. Extensive simulations and real-world flight tests validate the effectiveness and generalizability of ACEMSL. The project is available at \url{https://github.com/NTU-UAVG/CTS-visual-drone-swarm.git}.    
\end{abstract}
}

\ifCLASSOPTIONpeerreview
\begin{center} \bfseries EDICS Category: 3-BBND
\end{center}
\fi
%
\IEEEpeerreviewmaketitle

\section{INTRODUCTION}
As microelectronic technology, sensors, and onboard computing capabilities have {\color{black}advanced significantly} in recent years, autonomous {\color{black}microdrones} have been widely used in various missions \cite{alotaibi2019lsar
}. The use of autonomous agents in target search is becoming increasingly necessary in smart warehouse systems and disaster management \cite{chen2021integrated}, such as parcel pickup and delivery, and building collapse rescue. These scenarios demand efficient environment exploration, accurate situation awareness, and effective navigation in cluttered environments. However, the limited sensing range of a single drone can hinder the success of these tasks. {\color{black}Using a drone swarm, on the other hand, allows more comprehensive observation information and the ability to execute tasks in parallel \cite{chen20223}, making it an ideal solution for efficient target search. A drone swarm with formation maneuverability \cite{10040975} has been investigated to perform primary inspection tasks in desired geometric shapes but lacks significant collaborations. However, effective collaboration among agents is crucial for a drone swarm to successfully carry out the aforementioned tasks. It allows drones to assist each other in increasing the success rate of tasks and reducing the time and effort required to maneuver in cluttered environments without collisions.}

\begin{figure}[!tbp]
      \centering
      \includegraphics[width=3in]{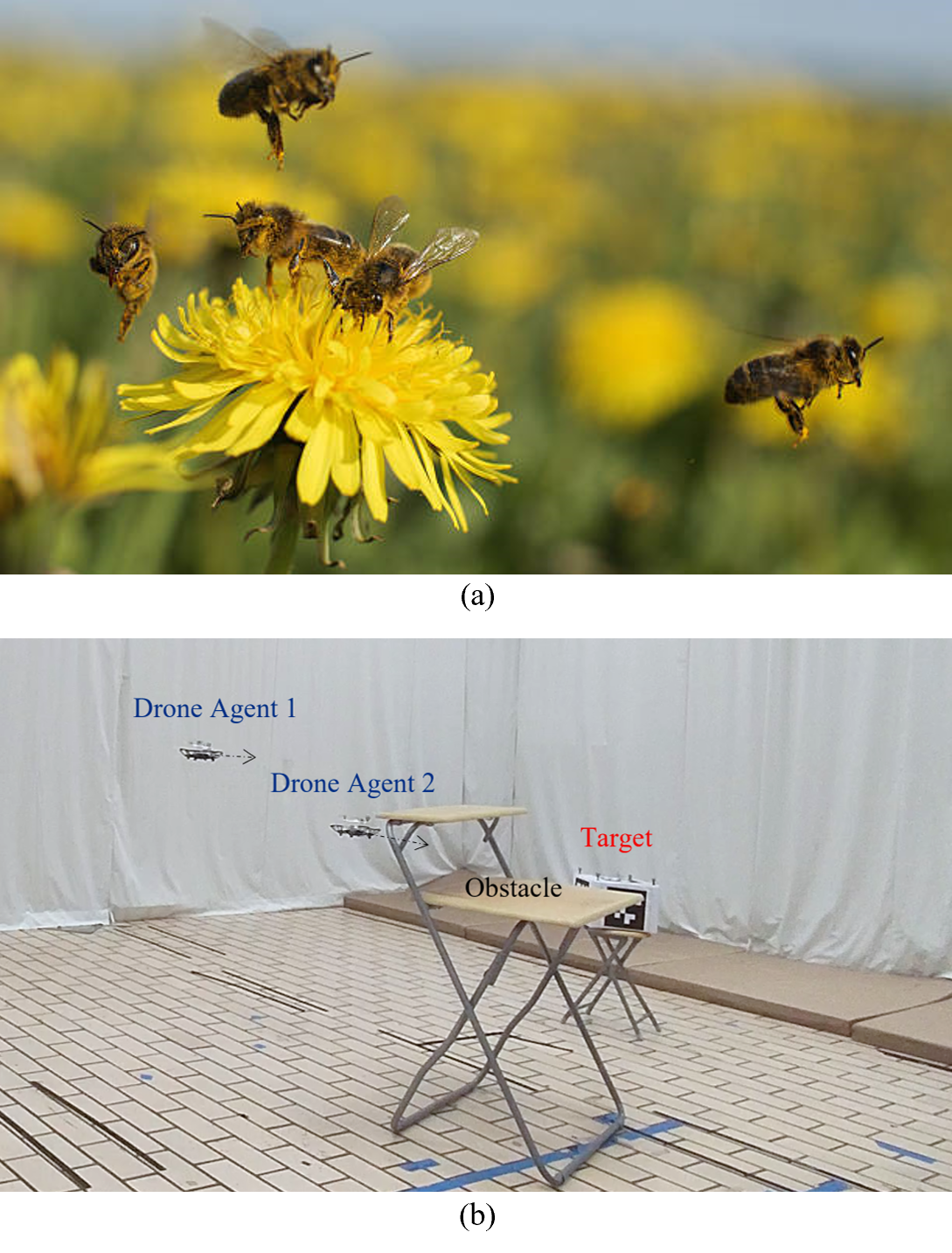}
      \caption{(a) A honeybee swarm is searching for nectar collaboratively [https://www.gettyimages.com/photos/bee]. (b) A visual drone swarm is searching for a target parcel box controlled by reinforcement learning.}
      \label{honeybees}
\end{figure}

Collaborative Target Search (CTS) behavior is commonly observed among intelligent biological swarms \cite{spaethe2006visual}, such as honeybees searching for nectar (see Fig. \ref{honeybees}), ants finding food and birds looking for a living space. {\color{black}CTS tries to locate and reach the target using multiple agents while efficiently exploring the environment. Achieving effective collaboration in a drone swarm is rather challenging, as the individual agent must be able to make intelligent decisions without the guidance of a leader or knowledge of the global environment map. To accomplish CTS, real-time perception and coordinated decision-making and planning are fundamentally required \cite{queralta2020collaborative}.}

The conventional idea to accomplish the CTS task is to formulate an optimization problem and break it down into local optimization sub-problems. There are two traditional ways to solve this local optimization problem, namely numerical analysis methods \cite{Haumann2010,Soria,8460996,8794090} and {\color{black}heuristic methods \cite{cao2015multi,7084641,hayat2017multi,zheng2019evolutionary,duisterhof2021sniffy, 9247966, 9210735}.} Under the category of numerical analysis methods, Haumann et al. \cite{Haumann2010} achieved multi-robot exploration and path planning via optimization partitioning of the objective function for each agent. Soria et al. \cite{Soria} adopted Nonlinear Model Predictive Control (NMPC) to optimize the navigation performance of an aerial robotic swarm towards a known targeted area in a cluttered indoor environment. The numerical analysis method necessitates knowledge of precise agents' dynamics and complex computation to obtain feasible solutions that satisfy constraints and boundary conditions. Therefore, it is extremely challenging to extend these methods to a large-scale swarm. Under the category of heuristic methods, based on the genetic algorithm, Hayat et al. \cite{hayat2017multi} formulated a multi-objective optimization algorithm to allocate tasks and find a target in a bounded area for a team of UAVs. In \cite{zheng2019evolutionary}, evolutionary algorithms (EAs) were developed to tackle criminal search problems with human-UAV collaboration. {\color{black}A search strategy based on particle swarm optimization (PSO) was proposed in \cite{duisterhof2021sniffy} for a nano-drone swarm to locate a gas source in unknown and cluttered environments.} These heuristic methods, on the other hand, can save modeling and computation resources but do not {\color{black}guarantee to obtain} the optimal solution for CTS tasks, resulting in feasible-only or local search behaviors {\color{black}even with perfect information}.

Recently, deep reinforcement learning (DRL) has been applied to robots for collaborative exploration \cite{8929168,9244647,9714163} in unknown environments. {\color{black}Luo et al. \cite{8929168} formulated the environment exploration problem in a graph structure and applied the graph convolutional network (GCN) to achieve space allocation. In \cite{9244647}, a hierarchical DRL integrated control architecture with dynamic Voronoi partitions was proposed to accomplish multiple mobile robots' cooperative exploration while relative target position observation was required. To address the problem of an exponential increase in the joint action space of multiple agents, Liu et al. \cite{9714163} proposed Feudal Latent-space Exploration (FLE) to improve the performance of multi-agent coordination.} However, differentiating from the full environment exploration problem, {\color{black} the CTS problem aims to infer the potential locations of an unknown target instead of maximizing the exploration area covered. Even though full environment exploration guarantees a feasible solution for target search, it would be more efficient for agents to directly focus on the potential space containing the target, i.e., probabilistic search methods \cite{8460996}.} {\color{black}Furthermore, real-time visual perception is challenging to integrate into coordinate exploration as it requires complex computation, which is rarely considered in existing DRL-based environment exploration methods.} While the target-driven visual navigation \cite{zhu2017target,9785445} with {\color{black}DRL} utilizes purely visual perception and a policy neural network to guide a robot through an indoor environment where the target is fixed within a spatial layout, {\color{black}our work focuses on the highly challenging multi-agent CTS mission over a 3D sparse reward space with only limited visual perception and egocentric observations and the target is randomly placed without prior knowledge. }

\textit{Contributions:}
{\color{black} To the best of our knowledge, this work is the first visual drone swarm adopted for CTS in a 3D cluttered environment to search for and approach a parcel box without prior target information.  The Success Rate (SR) and the Time To Reach (TTR) are the main metrics used to evaluate the performance of CTS.}

There are several challenges to {\color{black}accomplish} this CTS task for drones. {\color{black}(1)} \textbf{Firstly}, drone agents need to be motivated to efficiently explore the 3D environment with sparse rewards. {\color{black}(2)} \textbf{Secondly}, the drone is required to differentiate the target, obstacles, and other agents with {\color{black}{only}} forward color cameras during flight. {\color{black}(3)} \textbf{Lastly}, the drone swarm should display collaborative behaviors for target search, i.e., without colliding with obstacles or other agents. {\color{black}When it comes to real-world applications with limited sensing capabilities, these challenges become more pronounced and existing methods fail to address them effectively.}

To this end, we propose {\color{black} a novel and data-efficient DRL-based method named} Adaptive Curriculum Embedded Multistage Learning (ACEMSL) to address the aforementioned challenges. Compared to the standard curriculum learning methods \cite{bengio2009curriculum, xiao2021flying,damani2021primal} where the learning curriculum involves pre-designed parameters that change across the training process, our approach divides the training process into two adaptive curriculum embedded stages. This can reduce the amount of data required to achieve the best performance due to the self-adjustment capability of the curriculum. Meanwhile, CM3 (Cooperative Multi-goal, Multi-stage, Multi-agent) \cite{Yang2018} has validated multistage learning's data efficiency and outstanding performance for cooperative multi-goal multi-agent tasks like cooperative navigation. Similar to CM3, in our approach, a single greedy agent is trained prior to multiple agents' collaborative learning. However, CM3 considers its two stages as a complete curriculum, which is hard to extend {\color{black}on a} large scale and requires knowledge of goal assignment, while our ACEMSL embeds an adaptive curriculum into each stage without knowing the target's location. In each stage, the probability of hiding the target from a drone swarm increases when the desired {\color{black}SR} is achieved. {\color{black}As such, the drone agents are gradually guided to explore the environment with sparse rewards and try to search for the potential target.} To improve learning efficiency, the shared neural network of each agent is trained in a small space before being transferred to an unknown larger space. {\color{black}To enhance the generalization capability of the model,} domain randomization, which includes the initial state of drones, parameters of the target, and the intensity of light, is adopted during the training process. To validate our algorithm and learning framework, physical experiments with real-time visual perception are conducted in various scenarios. {\color{black}{Our \textbf{main contributions} are summarized as follows:}} 

{\color{black}1) Without any prior knowledge of the target and the environment layout, we propose a novel multistage reinforcement learning approach named ACEMSL to efficiently train a scalable decentralized visual drone swarm to collaboratively find the target with limited observation.

2) Differentiating from the pre-designed curriculum in the existing DRL-based environment exploration methods, we propose a novel adaptive embedded curriculum (AEC) algorithm to guide the visual drone to explore and navigate 3D cluttered environments with sparse rewards. The convergence of AEC is further proved.

3) To the best of our knowledge, our work is {\color{black}the first to} successfully transfer a trained policy from simulation to a real-life visual drone swarm and demonstrate their CTS performance with {\color{black}real-time visual perception} in physical experiments.}

{\color{black}\textit{Organization:}
The rest of this paper is organized as follows: Section II discusses existing DRL research on drone applications and multi-agent systems. Section III provides the basic preliminaries used to support our approach. Section IV describes the details of our proposed approach and algorithm. Section V presents the simulation experiments and physical experiments in detail, together with the performance comparison and analysis among different baseline methods. Lastly, Section VI concludes this paper.} 

\section{Related Works} \label{section 2}

In the domain of drones \cite{wu2019uav, xiao2021flying,song2021autonomous,Loquercio2021,9409767,o2022neural} and {\color{black}multi-agent system (MAS) \cite{Yang2018, berner2019dota, sartoretti2019primal, liu2020reinforcement, damani2021primal, 9404328}, DRL has achieved promising success}. The feasibility and transfer capabilities of DRL make it possible to address the collaborative exploration and target search problem in large-scale cluttered environments effectively. {\color{black}Existing DRL-based methods} in drone scenarios are mostly applied to single drone tasks \cite{song2021autonomous,xiao2021flying,Loquercio2021,o2022neural} or scenarios with discrete action spaces and grid observation environments \cite{Yang2018, damani2021primal}. In \cite{xiao2021flying}, an augmented curriculum learning approach was proposed to guide a drone through a narrow gap, but the pre-designed rules used to vary the curriculum limited the {\color{black}potential benefits of DRL.} {\color{black}Wu et al. \cite{wu2019uav} proposed a Snake algorithm with DRL to search for a target using an autonomous UAV and a global grid map, but failed to extend it to practical scenarios where only local visual observations are available.}

In \cite{Yang2018}, CM3 was proposed for multi-agent reinforcement learning (MARL) in {\color{black}a 2D space} to solve a collaborative navigation problem where the positions of landmarks and goals were known to each agent. Besides, function augmentation was adopted in CM3 to bridge the agent's own state in Stage 1 and the egocentric observations in Stage 2. However, this {\color{black} mechanism} hinders the scalability of the trained model, as a new model must be trained if the number of agents changes. \cite{liu2020reinforcement} proposed a two-level reinforcement learning-based control method for a UAV swarm to {\color{black}collaboratively} perform surveillance in an unknown 3D urban area. However, this approach requires that each UAV has information on the target's position and requires the environment to be constructed in grids {\color{black}with blocks of regular shape} to enable the agents to differentiate the obstacles and the target easily. This limits the application of this method in {\color{black}real-world scenarios}.

However, exploring a 3D unstructured environment and searching {\color{black}for} a hidden target with a visual drone swarm is more challenging since the 3D reward space is sparse and more constraints must be taken into consideration, such as continuous actions, limited observations and the scalability of the policy model. Our work addressed this challenging CTS problem for a visual drone swarm with only local visual perception, individual state, and limited egocentric observations. {\color{black}We improve scalability in our policy model by introducing a relative position measurement with the shortest distance as egocentric observation in multistage learning.}




\section{Preliminaries}
\subsection{Quadrotor Drone Dynamics}
The translational and rotational dynamics of a 6 degree-of-freedom (6DOF) quadrotor of mass $m$ can be described as follows, regardless of wind disturbance and aerodynamic drag:
\begin{subequations}
\setlength{\arraycolsep}{0.0em}
\begin{eqnarray}
   {\bm{\dot {p}}_{W}}&{}={}&\bm{v}_{W}\\
   {\bm{\dot {v}}_{W}}&{}={}& \bm{R}_B^W(\bm{q}_{WB}) \bm{f} + \bm{g}\\
   {\bm{\dot q}_{WB}}&{}={}& \frac{1}{2}\bm{\Omega}(\bm{\omega}_B)\bm{q}_{WB}
\end{eqnarray}
\setlength{\arraycolsep}{5pt}
\end{subequations}
{\color{black}where, ${\boldsymbol{p}_{W} = [x,y,z]^T}$ and ${\boldsymbol{v}_{W} = [v_x, v_y, v_z]^T}$ are the position and translational velocity vectors of the quadrotor in the world inertial frame $O_W$, where the $z$ axis of the world frame is aligned with the direction of the gravity $\boldsymbol{g}$. $\boldsymbol{q}_{WB} = [q{_0},q_1,q_2,q_3]^T$ denotes a unit quaternion which is used to represent the {\color{black}quadrotor's} attitude, while ${\boldsymbol{\omega}}_B = [\omega_x,\omega_y,\omega_z]^T$ denotes the body angular velocity (roll, pitch, and yaw, respectively) in the body frame $O_B$. $\bm{R}_B^W$ defines the transform matrix from the body frame $O_B$ to the world inertial frame $O_W$, and it is a function of $\boldsymbol{q}_{WB}$. $\boldsymbol{f}=[0,0,f]^T$ is the mass-normalized thrust vector with $f = \sum_{i=1}^{4} f_i^B/m$ {\color{black}(where $f_i^B$ are the trust from four propellers)} and $\boldsymbol{g} = [0,0,-g_z]^T$ with $g_z = 9.81m/s^2$ being the gravitational acceleration on Earth.} We denote $\boldsymbol{\Omega}(\boldsymbol{\omega}_B)$ as the skew-symmetric matrix given by
\begin{equation} \label{skew-sym}
\bm{\Omega}(\bm{\omega}_B) = \left[ {\begin{array}{*{20}{c}}
0&{ - {\omega _x}}&{ - {\omega _y}}&{ - {\omega _z}}\\
{{\omega _x}}&0&{{\omega _z}}&{ - {\omega _y}}\\
{{\omega _y}}&{ - {\omega _z}}&0&{{\omega _x}}\\
{{\omega _z}}&{{\omega _y}}&{ - {\omega _x}}&0
\end{array}} \right]
\end{equation}
$\bm{x} = [\boldsymbol{p}^T_{W}, \boldsymbol{v}^T_{W}, \boldsymbol{q}^T_{WB}]^T$ and $\boldsymbol{u} = [f, \omega_x,\omega_y,\omega_z]^T$ are selected as the states and the control inputs of a quadrotor, respectively. {\color{black}The body's momentum is not examined because the deployed high-bandwidth controller can precisely track the angular velocity commands,} allowing angular dynamics to be ignored.

\subsection{Reinforcement Learning}
Reinforcement Learning (RL) is a technique for mapping state space into the action space in order to maximize a long-term return with given rewards. The learner is not explicitly told what action to carry out, but must figure out which action will yield the highest reward during the exploring process. 
\subsubsection{Single-Agent RL}
{\color{black} The RL problem can be modeled with a Markov Decision Process (MDP) \cite{sutton2018reinforcement}, which is defined by a tuple $\langle \sset, \mathcal{O}, \aset, \Tfun, \rset, \D \rangle$ where $\sset$ denotes a set of possible states of agents. At time step $t$ and current state $s_t \in \sset$, the agent obtains an observation $o_t \in \mathcal{O} \subseteq \sset$ and selects an action $a_t \in \aset$ guided by a stochastic policy $\p(a_t \mid o_t) : \mathcal{O} \times \aset \mapsto [0,1]$. $\Tfun(s_{t+1}\mid s_t,{a}_t): {\sset \times \aset \times \sset \mapsto [0,1]}$ is the transition probability of the environment moving from the current state $s_t$ to the next state $s_{t+1}$ after taking an action ${a}_t$. $\rset : \sset \times \aset \mapsto \mathbb{R}$ refers to the reward function evaluating the bounded instantaneous reward {\color{black}$r_t := \rset(s_t, a_t) \in [\rmin, \rmax]$ for agent taking action $a_t$ under state $s_t$.} The agent starts from the state $s_0$ sampled from an initial state distribution $s_0 \sim \issetdef$. Given a policy $\pi$ and a state $s_t$, the state value function of a finite MDP $V : \sset \mapsto \rset$ is defined as:
\begin{equation}
    V_{\pi}(s_t) := \mathop{\mathbb{E}_{\pi}\left[ \sum_{k = 0}^{k = T-t}{\D}^{k}{r_{t+k}\mid{s_t}}\right]}
\end{equation}
where $\Ddef$ is the discount factor and $T$ is the final step of an episode. Similarly, considering the value of taking an action $a_t$, the state-action value function (Q-function) $Q : \sset \times \aset \mapsto \rset$ can be calculated as a Bellman function:
\begin{subequations}
\setlength{\arraycolsep}{0.2em}
\begin{eqnarray}
    Q_{\pi}(s_t, a_t) &:=& \mathop{\mathbb{E}_{\pi}\left[ \sum_{k = 0}^{k = T-t}{\D}^{k}{r_{t+k}\mid{s_t, a_t}}\right]}\\
    & = & \mathop{\mathbb{E}_{s_{t+1}\sim \Tfun(\cdot\mid{s_t, a_t})}\left[ r_t + \D V_{\pi}(s_{t+1})\right]}\\
                    & = &  \mathop{\mathbb{E}_{s_{t+1}\sim \Tfun(\cdot\mid{s_t, a_t})}\left[r_t + \D \mathop{\mathbb{E}_{a\sim \pi}\left[ Q_{\pi}(s_{t+1}, a_{t+1})\right]}\right]} {\nonumber}\\
\end{eqnarray}
\setlength{\arraycolsep}{5pt}
\end{subequations}

In policy-based methods, the objective of the RL is to directly find an optimal policy $\pi_{\theta}^* : \mathcal{O} \mapsto \aset$ to maximize the cumulative reward along a state-action trajectory $\tau := \{s_0, a_0, s_1, a_1, ..., s_T, a_T\}$ by adjusting the weights $\theta$ of the parameterized policy $\p_{\theta}$ over finite time steps (a training episode). The objective function can be formulated as
\begin{equation}
{\mathcal{J}({\p}_{\theta})} = \mathop{\mathbb{E}_{{s_0 \sim d_0},{\tau}\sim \pi_\theta}\left[ \sum_{t = 0}^{t = T}{\D}^{t}{r_{t}\mid{s_0}}\right]}
\end{equation}
The optimal policy is therefore obtained by maximizing the state value function $V_{\pi}(s_0)$ starting from $s_0$, i.e., 
\begin{equation}
{{{\p}}^{*}_{\theta}} = \argmax_{\theta}V_{\pi_\theta^{*}}(s_0)
\end{equation}

The optimal policies also share the same optimal Q-function, i.e., ${{{\p}}^{*}_{\theta}} = \argmax_{\theta}Q_{\pi_\theta^{*}}(s_0,a_0)$. The optimal Q-function satisfies the Bellman optimality equation:
\begin{equation}
    Q_{\pi^*}(s_t, a_t) = \mathop{\mathbb{E}_{s_{t+1}\sim \Tfun(\cdot\mid{s_t, a_t})}\left[ r_t +\D \max_{a_{t+1}} Q_{\pi^*}(s_{t+1}, a_{t+1})\right]}
\end{equation}

In the actor-critic (AC) framework\cite{konda1999actor}, the policy network $\pi_\theta(a_t \mid o_t)$ is the actor network and the Q-function $Q_{\pi_\theta}(s_t, a_t)$ is the critic network. The optimal policy $\pi_\theta^*$ can be obtained from policy iteration along the gradient of the critic network. The policy gradient theorem \cite{sutton1999policy} with learning rate $\alpha_t$ is:
\begin{equation}
    \nabla_\theta{\mathcal{J}({\p}_{\theta})} = \mathop{\mathbb{E}_{{s_t \sim \Tfun },{a_t}\sim \pi_\theta}\left[\nabla_\theta \log \pi_\theta(a_t \mid o_t)  Q_{\pi_\theta}(s_t, a_t)\right]}
\end{equation}
\begin{equation} \label{sga}
    \theta_{t+1} = \theta_t + \eta_t \nabla_{\theta_t}{\mathcal{J}({\p}_{\theta_t})}
\end{equation}
}
\subsubsection{Multi-Agent RL}
Similarly, the MARL problem can be formulated as a Multi-Agent Finite Markov Decision Process (MAFMDP) \cite{sutton2018reinforcement}. {\color{black} Similar to the single-agent RL, the MAFMDP is defined with a joint tuple $\langle \sset, \{\mathcal{O}^n\}, \{\aset^n\}, \mathcal{G}, \Tfun, \{\rset^n\}, N, \D \rangle$ with $N$ agents denoted by $n\in [N]$}. Since there is no knowledge of the target in our work, goal signals $\mathcal{G}$ can be ignored in the following description. Each agent $n$ obtains a {\color{black}local} observation $o_t^n := o^n(s_t) \in \mathcal{O}^n$ from the global state $s_t \in \sset$, and {\color{black}takes} an action $a_t^n \in \aset^n$. {\color{black}Let $\aset := {\aset}^1 \times \dots \times {\aset}^N$, the transition probability $\Tfun(s_{t+1}\mid s_t,\boldsymbol{a}_t) : {\sset \times \aset \times \sset \mapsto [0,1]}$ can be formulated with joint action $\boldsymbol{a}_t := \{a^1_t, ..., a^N_t\}$. For each agent $n$, the joint action can also be described as $\boldsymbol{a}_t := \{a^n_t, a^{-n}_t\}$ where $a_t^{-n}$ denotes all other agents'(except $n$) actions.} Each agent will receive an instantaneous reward $\rset^n_t := \rset(s_t, {a}^n_t)$ at each time step. {\color{black}Note that the instantaneous rewards include the team reward based on multi-agent credit assignment \cite{NIPS2003_c8067ad1}.} A joint stochastic policy is denoted as $\boldsymbol{\p}({\boldsymbol{a}_t}\mid{s_t}) := \prod_{n=1}^N \p^n(a^n_t \mid o^n_t)$ for all agents {\color{black}transiting} from the current state $s_t$ to the next state $s_{t+1}$ with the probability $\Tfun(s_{t+1}\mid s_t,\boldsymbol{a}_t)$ {\color{black}where $\p^n(a^n_t \mid o^n_t) : \mathcal{O}^n \times \aset^n \mapsto [0,1]$ is the decentralized stochastic policy for the agent $n$.} The objective of MARL is to find a joint policy to maximize cumulative reward along the action trajectory $\tau^n$. The objective function of MARL is:
\begin{equation} \label{marl-obj}
{\mathcal{J}({\boldsymbol{\p}}_{\pparams})} := \mathop{\mathbb{E}_{{\tau^n}\sim\boldsymbol{\pi}}\left[ \sum_{t = 0}^{t = T}{\D}^{t}\sum_{n = 1}^{n = N}{\rset^n_{t}\mid{s_0}}\right]}
\end{equation}

The joint optimal policy is obtained by:
\begin{equation} \label{opt-the}
{{\boldsymbol{\p}}^{*}_{\pparams}} = \argmax_{\pparams}{\mathcal{J}({\boldsymbol{\p}}_{\pparams})}
\end{equation}

The optimal policies for each agent are achieved along the policy gradient with a centralized critic $Q_{\boldsymbol{\p}}(s_0,\boldsymbol{a}_0)$ (for fully cooperative teams, where $\rset^1=\cdots=\rset^N$) \cite{foerster2018counterfactual} or their critics $Q_{\boldsymbol{\p}}^n(s_0,\boldsymbol{a}_0):=\mathop{\mathbb{E}_{\boldsymbol{\pi}}\left[ \sum_{t = 0}^{t = T}{\D}^{t}{\rset^n_{t}\mid{s_0, \boldsymbol{a}_0}}\right]}$ with {\color{black}a shared policy conditional on joint action \cite{Yang2018}.}

\section{Methodology}
\subsection{Problem Formulation}
In this section, the CTS problem is formulated into a MARL problem. The agent $n$ in this paper refers to the visual drone. Our goal is to learn {\color{black} a {\color{black}scalable} decentralized optimal shared policy $\pi^{*}_{\theta}: {\boldsymbol{\p}}^{*}_{\pparams} = \prod_{n=1}^N {\pi^{n,*}_{\theta}}$} for a drone swarm (a fully cooperative team) with $N$ agents to quickly find and navigate toward a hidden target denoted by $g$ and avoid collisions (can be formulated by rewards $\rset$){\color{black}. Only raw color images, ego states}, and egocentric measurements are available in the {\color{black}observation space} $\mathcal{O}^n$. There is no prior knowledge of the target $g$, and the only connection between each drone is the range sensor, from which the drone can obtain the relative positions of other agents. During the inference process, each agent continues to take actions based on the decentralized optimal policy until the termination condition is triggered.

\subsubsection{Environment Setup}
\begin{figure}[!tbp]
      \centering
      \includegraphics[width=2.5in]{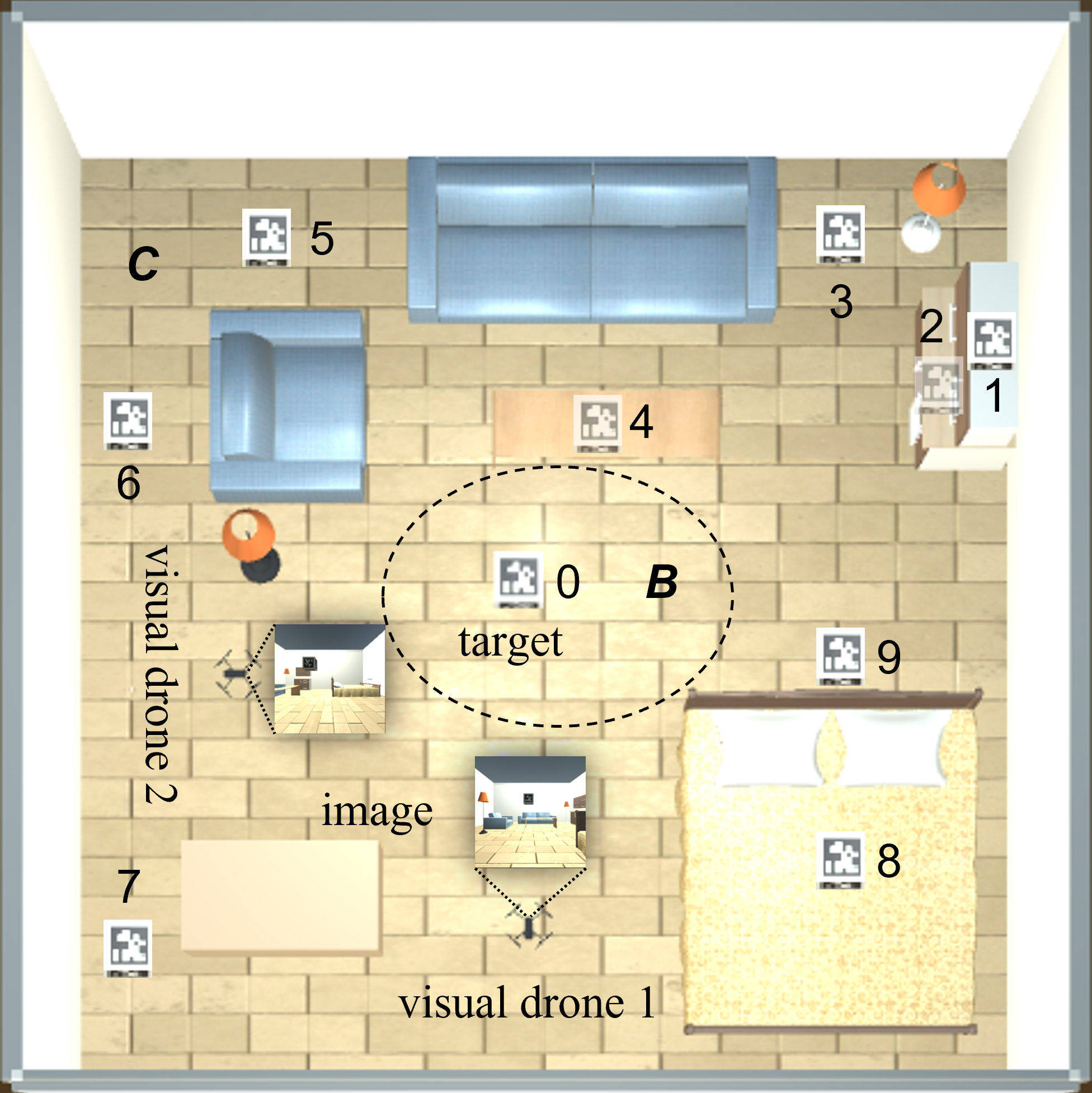}
      \caption{A bird's-eye view of the base environment, which consists of visual drone agents, obstacles, and a target. The furniture objects are the obstacles, such as sofas, the bed, and tables; The cubic textured with AprilTag is the target that drone agents need to find and reach. Each object is bounded by a collider for collision detection. Target can spawn in a closed sphere set $\boldsymbol{B}$ which is easily seen, or a discrete set $\boldsymbol{C}$ consists of several hidden locations marked with numbers $1-9$.}
      \label{figure1}
\end{figure}

To train and evaluate agents, a high-fidelity simulation environment is required in which we can add customized objects, such as obstacles, agents, and sensors, to represent the global states $\sset$. We build our simulation environment using Unity rendering engine\footnote{https://unity.com/} which allows us to generalize our model into different scenarios. The simulation scenarios become increasingly complex over different training stages. However, the simulation's base environment, which contains the agent drones and objects, is largely unchanged. Fig. \ref{figure1} shows the base environment, which consists of drone agents, obstacles, and a box target. More complex environments can be generated by {\color{black}increasing the size of the room, adding more obstacles, and hiding the target} at {\color{black}corners} where the drone agents can hardly find. 

The target can be spawned randomly within a closed sphere set $\boldsymbol{B}(ct, rd)$, in which the drone can easily detect at first sight. $ct$ and $rd$ are the center point and the radius of the sphere, respectively. Otherwise, it will be randomly placed at a predefined hidden corner drawn from a discrete set $\boldsymbol{C}(n_{loc})$. $n_{loc}$ is the size of $\boldsymbol{C}$, i.e., the number of hidden locations chosen. For instance, we can choose these locations with $n_{loc}=5$, namely, under the desk, behind the couch, below the cupboard, below the lamp, and above the cupboard. The probability of spawning a target at different locations is controlled by the spawning probability threshold $\epsilon \in [0,1]$. A random value $r_{seed}$ within {\color{black}the} interval $[0,1)$ is generated when each episode begins. If $r_{seed}$ is greater than $\epsilon$, the target will be spawned in $\boldsymbol{B}$, otherwise, it will be hidden at a place of $\boldsymbol{C}$. The probability density function (pdf) of spawning a target can be described as:
\begin{equation}
\label{spawnpdf}
\setlength{\nulldelimiterspace}{0pt}
{Pr(target)} = \left\{ \begin{array}{l}
U(\boldsymbol{B}),\quad r_{seed} > \epsilon\\
U(\boldsymbol{C}), \quad r_{seed} \leq \epsilon;
\end{array} \right.   
\end{equation}
where $U(\cdot)$ is a uniform distribution among a set. The spawning probability threshold $\epsilon$ will be adaptively adjusted in a curriculum learning way in each stage to make the task more challenging for drone agents. The validity of spawning positions will be checked, where if the target's shape {\color{black}overlaps} with other objects, the position is regarded as invalid, and the target will be spawned again.

The components of each obstacle are outlined by Unity’s mesh collider, which allows for the detection of collisions with another collider. The drone agent is modeled as a visual quadrotor equipped with a monocular RGB camera in front of the drone. The {\color{black}raw image} of fixed pixel size obtained from the forward camera can be used for recognizing the target and navigating towards the target. If the drone agent collides with the obstacles, other agents, or the target, a penalty (negative reward value) will be imposed.

\subsubsection{Observation Space}
The agent must decide which action to take based on the current observations $o^n_t$ to maximize the rewards. The drone agent receives observation information from its onboard camera and state sensors, such as IMU and positioning device. The image observation, the {\color{black}egostate observation, and the egocentric observation are denoted as $o^n_{I,t}$, $o^n_{S,t}$ and  $o^n_{E,t}$, respectively.} The RGB {\color{black}raw image} from the camera is down-sampled to a fixed resolution $3\times224\times224$ image $I_t \in  {\color{black}\mathbb{R}^{3\times 224 \times 224}}$ at every time step. The agent $n$ knows its own translational and rotational states, i.e., local position $\boldsymbol{p}^n_{W}$ and rotational quaternion $\boldsymbol{q}^n_{WB}$, and the relative positions of other team agents $\boldsymbol{p}^{nj}_{W}$, where $j\in[N]^{-n}$. However, only the relative position with the shortest distance $\boldsymbol{p}^{nj^*}_{W}$, i.e., $\|\boldsymbol{p}^{nj^*}_{W}\| = min(\|\boldsymbol{p}^{nj}_{W}\|)$ for $j\in[N]^{-n}$, is considered as the input observation of the policy model. Besides, a normalized forward direction vector $\boldsymbol{\bar{d}}^{n}_{W} \in \mathbb{R}^{3}$ is included in the observation space to ensure that the drone agent senses the direction. To ensure smooth maneuvering for the drone, the last action $a^n_{t-1}\in \mathbb{R}^{4}$ is also stored as an observation.

For each drone, only the current RGB image $I^n_t$ {\color{black}is} used and encoded into a vector $o^n_{EV,t} \in \mathbb{R}^{512}$ with a Convolutional Neural Network (CNN). This CNN allows the drone to detect different objects in their environment, such as obstacles, team agents, and the target, and sense the depth information. The {\color{black} concatenated} observation vector for the drone $n$ at time step $t$ is $o^n_t=\left[o^n_{EV,t}, \boldsymbol{p}^n_{W,t},\boldsymbol{q}^n_{WB,t}, \boldsymbol{\bar{d}}^n_{W,t}, a^n_{t-1}, \boldsymbol{p}^{nj^*}_{W,t} \right]\in\mathbb{R}^{529}$.

\subsubsection{Action Space}
In this work, we only focus on the high-level controller for drones. The low-level controller inside the drone tries to keep track of the high-level commands via inputs $\boldsymbol{u}$. Since the quadrotor is an under-actuated system, the selected high-level commands (actions $a_t^n$) are the four velocity commands generally used in the Robot Operation System (ROS), i.e., the desired velocity of the quadrotor in body frame $O_B$, namely $\hat v_x^B$, $\hat v_y^B$, $\hat v_z^B$ and $\hat \omega_z$. The action space $\aset^n$ consists of these four continuous actions that drive the drone to fly towards the target while exploring the constrained 3D space.

\subsubsection{Reward Function}
The reward function consists of two parts: the existential penalty $\rset_P = -t/T_{max}$ of the team, where $T_{max}$ is the allowable time steps in an episode, and the terminal reward $\rset_T$. The existential penalty {\color{black}accelerates} the {\color{black}exploring progress and} the terminal reward serves to guide the drone to fly towards the target {\color{black}in} the correct forward direction and to avoid {\color{black}crashing} simultaneously.

The terminal reward is a set function described as:
\begin{equation} \label{terminalReward}
\setlength{\nulldelimiterspace}{0pt}
{\rset_T} = \left\{ \begin{array}{l}
{+5},\quad if\;agent\;reaches\; the\; target\;\\
r_C, \quad if\;agent\;crashes\;
\end{array} \right.   
\end{equation}
Where $r_C$ is the penalty when the drone {\color{black}collides} with the obstacles or other agents.
\begin{equation}
    r_C = - \alpha\frac{\Vert \boldsymbol{d}_T \Vert
    }{\Vert \boldsymbol{d}_{init} \Vert} -\frac{\beta}{\pi} \arccos{\frac{\langle\boldsymbol{d}_T, \boldsymbol{\bar{d}}_{W}\rangle}{\Vert \boldsymbol{d}_T\Vert \cdot  \Vert \boldsymbol{\bar{d}}_{W} \Vert}} - 3
\end{equation}
where $\boldsymbol{d}_T$ is the vector from the collision position to the target position and $\boldsymbol{d}_{init}$ is the vector from the initial position to the target position; $\Vert \cdot \Vert$ denotes the norm of the vector and $\langle\cdot\rangle$ is the inner product of two vectors; $\alpha \in (0,1]$ and $\beta \in (0,1]$ are the weights {\color{black}used} to adjust the penalty from the distance to fly and the penalty caused by the forward direction. Note that the terminal reward $r_C$ is accounted for each agent, and the episode terminates only when all agents crash. {\color{black}The positive reward of $+5$ is a team reward and distributed with an explicit \textbf{equal credit assignment}, i.e., each agent receives $+5$ if one agent reaches the target.} It is important to note that $\rset_P \geq -1$ and $min(\rset_T) < \rset_P < max(\rset_T)$ to prevent the agents from taking suicidal actions without motivation to complete the task. Additionally, we enforce the condition $-r_C \leq max(\rset_T)$ to encourage the agents to explore the environment instead of staying in a safe space. {\color{black} Reward scale is also an essential factor affecting the performance of DRL. According to the results from \cite{henderson2018deep}, a reward scale $\hat{\sigma} = 10$ is preferred since no layer normalization is used for our network. Hence, $\rset_T \in [-10, 10]$ is considered in this work. We further verify a larger value such as ``+10" does not change the performance significantly.}
\begin{figure}[!tbp]
      \centering
      \includegraphics[width=3.3in]{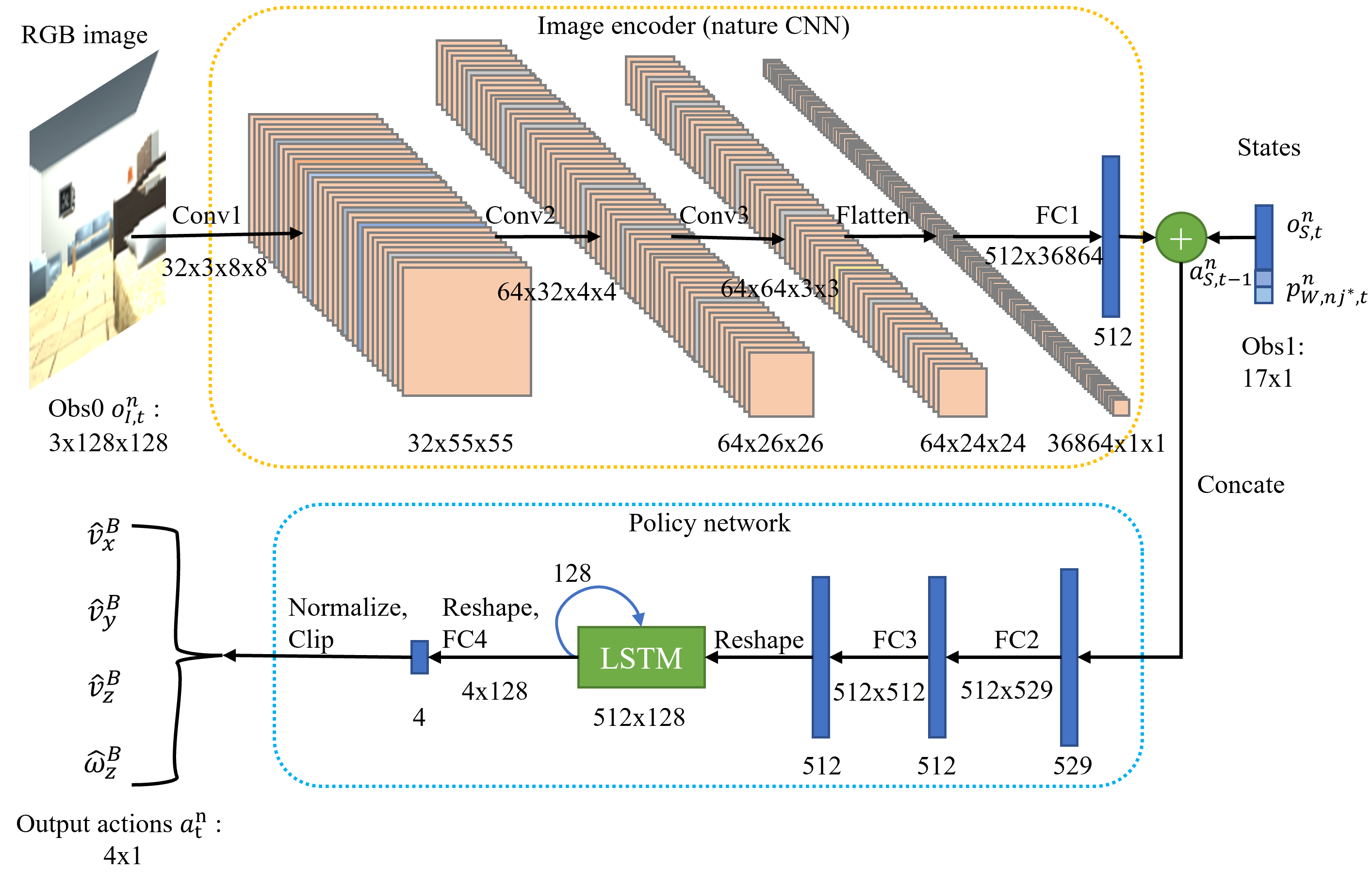}
      \caption{Illustration of the neural network architecture for target search, which consists of the image encoder, the policy network, and the memory network.}
      \label{figure2}
\end{figure}
{\color{black}
\begin{theorem}[Existence of Optimal Policy]
\label{existence} With multi-agent equal credit assignment, the optimal shared policy $\p^{*}(a^n_t \mid o^n_t)$ for CTS exists and can be achieved along the improvement of local critic $Q^n_{\boldsymbol{\p}}(s_0,a^n_0) := \mathop{\mathbb{E}_{\boldsymbol{\p}}\left[\sum_{t = 0}^{t = T}{\D}^{t}{\rset^n_t\mid{s_0, a^n_0}}\right]}$ conditional on the joint action $\textbf{a}_t$ and the local observations $\{o^n_t\}$.
\end{theorem}

\begin{proof}
With equal credit assignment, the CTS becomes a fully cooperative game, i.e., the reward function of each agent is the same $\rset^1=\cdots=\rset^N$. For each agent $n$, accumulate the joint probability over $a^{-n}$ and we have the Q-function as follows,
\begin{equation}
\begin{split}
    Q^n_{\boldsymbol{\p}}(s_0,a^n_0)
    &=\sum_{a^{-n}}\p(a^{-n}\mid s)Q_{\boldsymbol{\p}}^n(s_0,a^n_0, a^{-n}_0) \\
    &=\sum_{a^{-n}}\p(a^{-n}\mid s)Q_{\boldsymbol{\p}}^n(s_0,\boldsymbol{a}_0) 
\end{split}
\end{equation}

The objective function (\ref{marl-obj}) can then be described with
\begin{equation}
\begin{split}
    {\mathcal{J}({\boldsymbol{\p}}_{\pparams})} &= \sum_{n = 1}^{n = N}\mathop{\mathbb{E}_{\boldsymbol{\pi}}\left[\sum_{t = 0}^{t = T}{\D}^{t}{\rset^n_{t}\mid{s_0}}\right]} \\
    &=\sum_{n = 1}^{n = N}V^n_{\boldsymbol{\pi}}(s_0) \\
    &=\sum_{n = 1}^{n = N}\mathop{\mathbb{E}_{\boldsymbol{a}\sim\boldsymbol{\pi}(\boldsymbol{a}\mid s)}Q_{\boldsymbol{\p}}^n(s_0,\boldsymbol{a}_0)}\\
    &=\sum_{n = 1}^{n = N}\mathop{\mathbb{E}_{a^{n}\sim {\pi}({a^n}\mid o^n)}\sum_{a^{-n}}\p(a^{-n}\mid s)Q_{\boldsymbol{\p}}^n(s_0,\boldsymbol{a}_0)}\\
    &=\sum_{n = 1}^{n = N}\mathop{\mathbb{E}_{a^{n}\sim {\pi}({a^n}\mid o^n)}Q^n_{\boldsymbol{\p}}(s_0,a^n_0)}
\end{split}
\end{equation}
According to \textbf{Proposition 1} in \cite{Yang2018}, we have a Bellman function $ Q^n_{\boldsymbol{\p}}(s_t,a^n_t) = \mathop{\mathbb{E}_{\boldsymbol{\pi}}\left[\rset^n_t + \D Q^n_{\boldsymbol{\p}}(s_{t+1},a^n_{t+1})\right]}$. Hence, the objective function can be further expanded as
\begin{equation}
\begin{split}
{\mathcal{J}({\boldsymbol{\p}}_{\pparams})} &= \sum_{n = 1}^{n = N}\mathop{\mathbb{E}_{a^{n}\sim {\pi}({a^n}\mid o^n)}\left[\mathop{\mathbb{E}_{\boldsymbol{\pi}}\left[\rset^n_0 + \D Q^n_{\boldsymbol{\p}}(s_{1},a^n_{1})\right]}\right]} \\
&= \sum_{n = 1}^{n = N}\mathop{\mathbb{E}_{\boldsymbol{\pi}}\rset^n_0} + \sum_{n = 1}^{n = N}\mathop{\mathbb{E}_{a^{n}\sim {\pi}({a^n}\mid o^n)}\left[\mathop{\mathbb{E}_{\boldsymbol{\pi}}\left[\D Q^n_{\boldsymbol{\p}}(s_{1},a^n_{1})\right]}\right]} \\
&= N\mathop{\mathbb{E}_{\boldsymbol{\pi}}\rset^n_0} + \sum_{n = 1}^{n = N}\mathop{\mathbb{E}_{a^{n}\sim {\pi}({a^n}\mid o^n)}\left[\mathop{\mathbb{E}_{\boldsymbol{\pi}}\left[\D Q^n_{\boldsymbol{\p}}(s_{1},a^n_{1})\right]}\right]}
\end{split}
\end{equation}
Hence, at each episode $t$, we select the greedy policy $\pi'(a^n\mid o^n) := \argmax_a Q^n_{\boldsymbol{\p}}(s_{t},a^n_{t})$ for agent $n$ and the shared policy $\pi'$ is the optimal policy satisfying (\ref{opt-the}).
\end{proof}
}

\subsection{Network Architecture}
As shown in Fig. \ref{figure2}, the neural network used in this paper consists of three parts: the image encoder, the policy network, and the memory network. The memory network is embedded {\color{black}within} the policy network for processing temporal sequence observations. We have conducted performance analysis on 3 visual encoders: a simple 2-layer CNN, a nature CNN \cite{Mnih2015}, and a ResNet \cite{Espeholt2018}. After considering efficiency and search performance, a shallow neural network-nature CNN is used in our model. The image encoder-nature CNN consists of 3 convolutional layers, 1 flatten layer, and 1 fully connected (FC) layer. A Leaky Rectified Linear Unit (Leaky ReLU) is adopted as the activation function for all convolutional layers and the encoder's FC layer. The image encoder {\color{black}converts} the RGB image of fixed resolution into a $512$-dimensional flattened vector. The policy neural network is designed with $2$ FC ($512$-nodes with a Sigmoid activation function) layers which {\color{black}map} the concatenated observation vector $o^n_t \in \mathbb{R}^{529}$ into a continuous action vector $a^n_t \in \mathbb{R}^{4}$. The memory network-Long Short Term Memory (LSTM) is a recurrent neural network used to feed previous outputs back and generate smooth trajectories for drones.

\subsection{Adaptive Curriculum Embedded Multistage Learning}
\subsubsection{Adaptive Embedded Curriculum (AEC)} In the standard curriculum learning methods \cite{bengio2009curriculum,xiao2021flying,damani2021primal}, the curriculum is manually designed with a static, discrete parameter sequence in which the task difficulty level {\color{black}(TDL)} is progressively increased with a pre-designed mode. Even though the pre-designed curriculum parameter can be post-adjusted according to the training results, {\color{black}obtaining a satisfactory curriculum can be inconvenient and inefficient.} Hence, it is hard for agents to achieve their best performance. {\color{black}In contrast to standard curriculum learning methods,} {\color{black} our proposed AEC aims to explore agents' proficiency capability, where} the {\color{black}TDL} is adaptively adjusted {\color{black}at the beginning of each episode} with a changing rate {\color{black}$\delta_\epsilon > 0$} based on the {\color{black}SR} of finding the target achieved by drone agents. If the {\color{black}SR} rises to a high boundary $sr_h$, the {\color{black}TDL} increases. Correspondingly, if the {\color{black}SR} drops to a low boundary $sr_l$, the {\color{black}TDL} decreases. {\color{black}The principle behind is that the higher {\color{black}TDL} of environment will try to decrease the {\color{black}SR} of the team while the drone team is learning to obtain a higher {\color{black}SR}. As such, the trained policy of the drone team and the {\color{black}TDL} of environment will converge to an equilibrium with this AEC algorithm. This equilibrium provides the optimal policy for the drone swarm to search for the target.}

There are several {\color{black}environment} parameters that can be considered to derive the {\color{black}TDL}, such as the size of the space, the number of obstacles, the size of the target, and the spawning probability threshold ($\epsilon$) in (\ref{spawnpdf}). In our scenario, the {\color{black}TDL} is formulated by the spawning probability threshold since our main goal is to find the hidden target as fast as possible. {\color{black}Hence, in the following discussion, the TDL is denoted by $\epsilon$ and clipped within $[\epsilon_{min}, \epsilon_{max}]$.} Other {\color{black}environment} parameters can be randomly selected in the domain randomization to improve the generalization of the trained policy. {\color{black}In this paper, we provide two ways to calculate the success rate. One is the cumulative success rate $sr = (success~episodes) / (all~episodes)$; another is the moving success rate $sr = (success~episodes~in~a~sliding~window) / wl$, where $wl$ is the sliding window length. The AEC with a cumulative success rate is described in Algorithm \ref{alg::target} and the AEC with a moving success rate is presented in Algorithm \ref{alg::AEC-SW}.}  

\begin{algorithm}[ht]
\caption{Adaptive Embedded Curriculum (AEC)}
\label{alg::target}
\KwIn{An initial {\color{black}TDL} $\epsilon_0$, changing rate $\delta_\epsilon$, $sr_l$, $sr_h$} 
\KwOut{The spawning position $\mathbf{p}_{goal}$ of target}
On the start of training: \\
{Initialize $\boldsymbol{p}_{W,0}$, $\boldsymbol{v}_{W,0}$, $\boldsymbol{q}_{WB,0}$, $SR=0.0$, $successCount=0$, $episodeCount=0$} \;
{{\color{black}Construct the critic Q-function $Q^n_{\boldsymbol{\p_{\theta}}}(s_0,a^n_0)$;}}\\
{\color{black} \For{episode ~$= 0$ : maximum episodes}{
{On each episode begin:}\\
\If{$episodeCount \neq 0$}{$SR \gets successCount/episodeCount$\;}
$episodeCount \gets episodeCount + 1$\;
\If{$SR > sr_h$}{$\epsilon \gets max(\epsilon+\delta_\epsilon, {\color{black}\epsilon_{max})}$\;}
\If{$SR < sr_l$}{$\epsilon \gets min(\epsilon-\delta_\epsilon, {\color{black}\epsilon_{min})}$\;}
Spawn the target according to (\ref{spawnpdf})\;
{\color{black}\For{step $t=0:T$}{Conduct policy improvement, for $t = 0$ to $T$;\\
\If{one drone reaches the target}{$successCount \gets successCount+1$;\\
end the episode\;}
\If{all drones crash}{end the episode;\\
back to 4;}}
}
}
}
\end{algorithm}

{\color{black}
\begin{theorem}[Convergence Analysis]
\label{Convergence} If (i) the policy gradient algorithm used for policy improvement converges to the optimal policy {\color{black}in} any fixed TDL $\epsilon \in [\epsilon_{min}, \epsilon_{max}]$ and (ii) the success rate $sr$ is reciprocal correlated with $\epsilon$, then AEC converges to the optimal policy, and the TDL converges to $\epsilon^*$ for any $\epsilon_0 > 0$ and  $sr_h \geq sr_l$  with finite training steps $T'$.
\end{theorem}
\begin{proof}
WLOG, let assume $sr = k/\epsilon + c$, with $k>0, c\geq0$. We have $\epsilon \gets max(\epsilon+\delta_\epsilon, \epsilon_{max})$ when $sr > sr_h$, i.e., $k/\epsilon + c > sr_h$. That is, when $\epsilon < k/(sr_h-c)$, $\epsilon \gets max(\epsilon+\delta_\epsilon, \epsilon_{max})$. Therefore, $\forall \epsilon_0 < k/(sr_h-c), \exists{T'}$ s.t. $\epsilon^* = (\epsilon_0 + T'\delta_\epsilon) \geq k/(sr_h-c)$. Similarly, when $sr < sr_l$, i.e. $\epsilon > k/(sr_l-c)$, $\epsilon \gets min(\epsilon-\delta_\epsilon, \epsilon_{min})$. Hence, $\forall \epsilon_0 > k/(sr_l-c), \exists{T'}$ s.t. $\epsilon^* = (\epsilon_0 - T'\delta_\epsilon) \leq k/(sr_l-c)$. Combine together, $\forall \epsilon_0 > 0, \exists{T'}$ s.t. $\epsilon$ converges to $\epsilon^* \in [k/(sr_h-c), k/(sr_l-c)] \cap [\epsilon_{min}, \epsilon_{max}]$.

Since the policy converges to the optimal policy $\pi^*(a^n\mid o^n,\epsilon)$ given any $\epsilon$, according to the policy improvement theorem, the AEC converges to the optimal policy $\pi^*(a^n\mid o^n,\epsilon^*)$. This completes the proof.
\end{proof}

\begin{remark}
The converge episodes required {\color{black}depend} on the changing rate $\delta_\epsilon$ and the sliding window length $wl$. With a higher value of $\delta_\epsilon$ and a smaller $wl$, the convergence progress is more likely {\color{black}to fluctuate}.
\end{remark}
}

\begin{algorithm}[ht]
{\color{black}
\caption{AEC with Sliding Window (AEC-SW)}
\label{alg::AEC-SW}
\KwIn{An initial {\color{black}TDL} $\epsilon_0$, changing rate $\delta_\epsilon$, $sr_l$, $sr_h$} 
\KwOut{The spawning position $\mathbf{p}_{goal}$ of target}
On the start of training: \\
{Initialize $\boldsymbol{p}_{W,0}$, $\boldsymbol{v}_{W,0}$, $\boldsymbol{q}_{WB,0}$, $movingSR=0.0$, $slidingWindow=\textbf{0}_{[wl]}$, $episodeCount=0$} \;
{{\color{black}Construct the critic Q-function $Q^n_{\boldsymbol{\p_{\theta}}}(s_0,a^n_0)$;}}\\
{\color{black} \For{episode ~$= 0$ : maximum episodes}{
{On each episode begin:}\\
\If{$episodeCount \neq 0$}{$movingSR \gets mean(slidingWindow)$\;}
$index \gets episodeCount \mod wl$\;
$slidingWindow[index] \gets 0$\;
$episodeCount \gets episodeCount + 1$\;
\If{$movingSR > sr_h$}{$\epsilon \gets max(\epsilon+\delta_\epsilon, {\color{black}\epsilon_{max})}$\;}
\If{$movingSR < sr_l$}{$\epsilon \gets min(\epsilon-\delta_\epsilon, {\color{black}\epsilon_{min})}$\;}
Spawn the target according to (\ref{spawnpdf})\;
\For{step $t=0:T$}{Conduct policy improvement;\\
\If{one drone reaches the target}{$slidingWindow[index] \gets 1$;\\
end the episode\;}
\If{all drones crash}{end the episode;\\
back to 4;}
}}
}
}
\end{algorithm}

\subsubsection{Stage 1}
At Stage 1, the objective is to train a single drone to identify, track and fly towards a randomly spawned target according to Algorithm \ref{alg::target} or \ref{alg::AEC-SW}. During Stage 1, the drone explores the environment with the designed reward function. The MARL problem is reduced into a single agent ($N=1$) problem, and the group reward becomes the agent $n$'s reward. The optimal policy achieved by agent $n$ is the greedy policy \cite{Yang2018}. The drone will learn to find and approach the stationary target without crashing into obstacles. 

\subsubsection{Stage 2}
Once the drone has successfully learned the target-approaching behavior, another drone is added to the scene in Stage 2 for generating collaborative behavior. The trained model in Stage 1 is used to initialize the model in Stage 2. These two drones are further trained to search for the same target without {\color{black}colliding with} each other. Once the agents crash (collide with the obstacles or other agents), their status becomes inactive until the next episode. The episode ends when all agents {\color{black}crash} or one agent reaches the target.

\subsubsection{Training Algorithm} \label{ppo-sac}
The Proximal Policy Optimization (PPO) algorithm \cite{schulman2017proximal} is selected for our policy training {\color{black} according to Theorem \ref{Convergence} even though the state-of-the-art (SOTA) off-policy algorithm Soft Actor-Critic (SAC) \cite{haarnoja2018soft} performs more data-efficient than PPO over general RL tasks especially in static environments, such as the Humanoid benchmark task. SAC adopts an entropy maximization term into the actor-critic framework to trade off the exploit-exploration and to improve the data efficiency via experience replay. However, from the perspective of an individual agent, our CTS scenarios have non-stationary states such as the positions of teammates and the target are always changing. This can cause significant divergence issues in the experience replay mechanism utilized by SAC, as indicated in \cite{foerster2017stabilising} and the  \href{https://github.com/Unity-Technologies/ml-agents/blob/main/docs/ML-Agents-Overview.md#training-methods-environment-agnostic}{overview page} of Unity ML-Agents Toolkit \cite{juliani2018unity}. Thus, PPO is preferred for our application since it is more effective and stable for problems with dynamic environments by using {\color{black}a ``clipped”} surrogate objective and on-policy learning. We further compared the performance of SAC and PPO in our Stage 1 training in Section \ref{Exp} to validate the advantages of PPO in our application.}

In our training process, only the states of the agents and the {\color{black}raw image} are collected as observations, without the need for information on relative positions between the agents and the target, which is usually the case for the traditional numerical methods\cite{zheng2019evolutionary}. Only at the end of each episode are the distance values counted in the reward function. Meanwhile, the existential penalty is designed to encourage the agent to reach the target as fast as possible.

\subsection{Domain Randomization}
Domain randomization \cite{tobin2017domain} is an effective technique to improve the generalization capability of the model during training and hence increase the success rate of Sim2Real \cite{tobin2017domain} transfer to unseen scenarios. At the beginning of each episode, environment parameters and agents' states are randomly sampled from uniform distributions. In our work, the domain randomization used can be divided into (1) environment randomization that involves randomization of parameters, such as light intensity in Unity scenes $I_v$, {\color{black}the scale $\lambda_{target}$, and} the yaw angle $\psi_{target}$ of the target; (2) agent randomization that involves randomization of parameters, such as the initial states of agents. The environment randomization extends the trained model to accomplish the tasks in unseen scenarios while the agent randomization enables the agents to start with perturbations in their initial configurations.

\begin{table}[] 
\caption{Domain Randomization}
\begin{center}
\centering
\label{domainR}
\begin{tabular}{ l l} 
\toprule
\textbf{Environment Randomization} \\
\hline
     Light intensity $I_v$ &$U(0.2, 1.0)$ \\
     Scale of target $\lambda_{target}$ &$U(0.2m,0.3m)$  \\
     Yaw angle of target $\psi_{target}$ &$U(0^{\circ},360^{\circ})$ \\
     \hline
\textbf{Agent Randomization} \\
     \hline
     Position noise & $U(\boldsymbol{B}(ci, 0.2m))$ \\
     Noise in yaw angle of agent  & $U(-30^{\circ},30^{\circ})$ \\
\bottomrule
\end{tabular}
\end{center}
\end{table}

As listed in Table \ref{domainR}, the environment parameters are randomly sampled from a uniform distribution $U(min, max)$, while the agent randomization is generated by adding a uniformly distributed random noise $U(\cdot)$. Note that {\color{black}the agent position noise} is sampled from a uniformly distributed closed sphere set $\boldsymbol{B}(ci,ri)$, i.e., $U(\boldsymbol{B}(ci,ri))$, where $ci$ is the initial position of agents and $ri$ is the radius of sphere set. $U(min, max)$ and $U(\boldsymbol{B}(ci,ri))$ are generated via the $Random.Range(min, max)$ function and $Random.insideUnitSphere*ri$ operator in Unity.

\section{Experiments and Results} \label{Exp}
\begin{figure}[!tbp]
      \centering
      \includegraphics[width=3.3in]{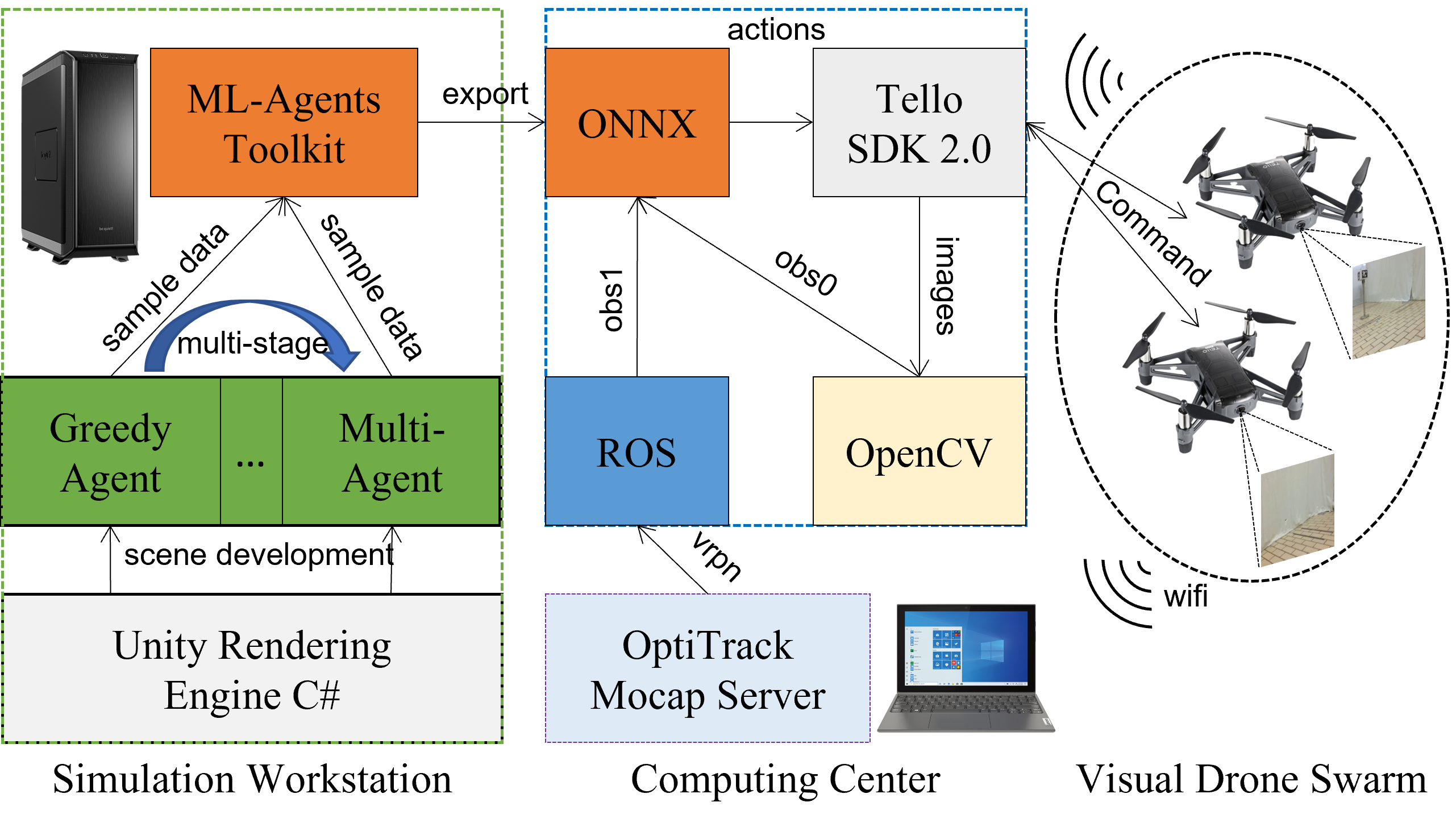}
      \caption{Illustration of the training and experiment framework.}
      \label{experiments_a}
\end{figure}

In this section, we present the simulation experiments, including the training process and inference tests, and the physical experiments covering the model Sim2Real transfer and the real-time flight tests. The architecture of simulation and physical experiment platform is illustrated in Fig. \ref{experiments_a}, which consists of a simulation workstation (with AMD Ryzen 9 5900X 12 cores CPU, Nvidia RTX 3090 Gaming OC 24GB GPU, and 32GB RAM), a laptop computing center, the OptiTrack motion capture server streaming the position data of selected rigid bodies and a Tello Edu visual drone swarm.

The simulation workstation is used to develop simulation environments for training and testing and export the trained neural network model for deployment. With the trained ONNX\footnote{https://onnx.ai/} model, the computing center acquires real-time observations from the motion capture server and Tello Edu drones, and generates corresponding actions which drive the Tello Edu visual drones to search for the target via Tello SDK 2.0\footnote{https://www.ryzerobotics.com/tello-edu/downloads}. The computing center is connected to the OptiTrack motion capture system to {\color{black}subscribe to} the position data of drones via Robots Operation System (ROS) nodes. The image observation is acquired from the onboard camera of Tello Edu drones using necessary image processing. Note that a Tello Edu drone can only be set as a wireless access point (AP) to stream the video. Hence, multiple USB WIFI adapters are used in the computing center for connecting to {\color{black}each drone separately using WIFI}.

\subsection{Simulation Experiments}
\subsubsection{Settings}Our simulation platform is developed based on the Unity rendering engine and ML-Agents Toolkit \cite{juliani2018unity}, a flexible simulation and training platform for multi-agent reinforcement learning. Scenes of target search with a single greedy agent and CTS with a drone swarm are developed successively. The dimensions of the base simulation environment in Fig. \ref{figure1} is $5m\times5m\times3m$ (width $\times$ length $\times$ height). In the AEC, the {\color{black}TDL} starts from $\epsilon_0=0.1$ and changes with $\delta_\epsilon =0.1$. {\color{black}The TDL is clipped within $[0.1,0.9]$, not its domain $[0.0,1.0]$ during the training process since we aim to avoid the situation where the policy converges to only searching the set $B$ at the beginning. The {\color{black}SR} boundaries are set to $sr_h = 0.85$ and $sr_l = 0.40$, which are the expected {\color{black}SR} and the allowable lowest {\color{black}SR} for our task, respectively.} The allowed maximum time step is $T_{max} = 5000$ in one episode. {\color{black} The reward weights are set to $\alpha = 1$ and $\beta = 0.1$ as we focus more on the distance to reach during training.} The initial linear and angular velocity of drones are reset to zero at the start of each episode. The parameters of AEC used in our simulation are concluded in Table \ref{ACEMSL-param}. 
{\color{black}These parameters are tried and adjusted in a reasonable range to reach the near-best performance.}

\begin{table}[]
{\color{black}
\centering
\caption{Parameters Used for the AEC}
\label{ACEMSL-param}
\resizebox{\columnwidth}{!}{
\begin{tabular}{@{}ll|ll@{}}
\toprule
\textbf{Parameter} & \textbf{Value}  &\textbf{Parameter} & \textbf{Value}\\
\hline

Environment size & $5\times5\times3 (m^3)$ & Initial TDL $\epsilon_0$ & 0.1 \\
Minimum TDL $\epsilon_{min}$ & 0.1 & Maximum TDL $\epsilon_{max}$ & 0.9 \\
TDL changing rate $\delta_\epsilon$ & 0.1 & High SR boundary $sr_h$ & 0.85 \\
Low SR boundary $sr_l$ & 0.40 & Maximum step $T_{max}$ & 5000 \\
Reward weight $\alpha $ & 1.0 & Reward weight $\beta$ & 0.1\\
Maximum speed $v_{max}$ & 1.0 m/s & Window length $wl$ & 1000\\
\bottomrule
\end{tabular}
}}
\end{table}

\subsubsection{Baselines and Ablations}
The performance of our approach is compared {\color{black}against SOTA methods: (1) TD-A3C \cite{zhu2017target}, which only uses visual observations; (2) CM3 \cite{Yang2018}, which is trained via two-stage learning but without AEC (with a fixed {\color{black}TDL} $\epsilon=0.3$).} We ensure that TD-A3C and CM3 are modified so that both can be applied for the target search tasks in the training. Since TD-A3C is proposed to address a single-agent navigation problem, we evaluate its performance with our PPO with AEC (PPO-AEC) and PPO without AEC (PPO-w/o-AEC) in the single-agent training. {\color{black} PPO-AEC and PPO-w/o-AEC are compared to examine the contribution of AEC to our approach. Meanwhile, PPO-AEC-SW is compared with PPO-AEC to investigate the effect of the sliding window scheme. To validate the advantages of PPO over SAC, PPO-AEC is compared with SAC-AEC (single agent trained with SAC algorithm) regarding the training runtime and task performance. Note that the SAC-AEC is computationally expensive; thus we only consider 3 million training episodes.} 

{\color{black}To investigate the data efficiency of the proposed multistage scheme, fully trained Stage 2 {\color{black}without pre-trained policy} (``Direct" method) is compared with CM3 and ACEMSL in the multi-agent training. {\color{black}Since we only compare the performance of ACEMSL and CM3 in Stage 2, Stage 1 of ACEMSL and CM3 are the same, when trained for $3$ million episodes with the AEC.} Note that for fair evaluation, we also trained TD-A3C and CM3 with the PPO algorithm since we consider the method frameworks, not the training algorithm itself. The hyperparameters of PPO for all training remain the same.} 

\begin{table}[]
\centering
\caption{Hyperparameters Used for Training}
\label{ppo-param}
{\color{black}
\resizebox{\columnwidth}{!}{
\begin{tabular}{@{}ll|ll@{}}
\toprule
\textbf{Parameter} & \textbf{Value}  &\textbf{Parameter} & \textbf{Value}\\
\hline

Batch size & 2048 & Buffer size & 10240 \\
Learning rate $\eta$ & 0.0003 & Discount factor $\gamma$ & 0.99\\
Number of epochs & 3 & Learning rate schedule & linear \\
Checkpoints & 10 & Time horizon & 128\\
\hline
\textbf{PPO} &  & \textbf{SAC}  & \\
Entropy bonus beta & 0.01 &Interpolation factor $\tau$ & 0.005\\
Clip threshold $\epsilon^{ppo}$ & 0.2 & Update steps & 10.0 \\
Regularization factor $\lambda^{ppo}$ & 0.95 &Initial entropy coefficient & 0.5\\ 
Optimizer & Adam & Replay size & 1000 \\
Maximum episodes & 12 million &Maximum episodes & 3 million\\
\bottomrule
\end{tabular}
}}
\end{table}

\subsubsection{Training}
{\color{black} The training hyperparameters are determined and fine-tuned based on the suggestions and results from \cite{henderson2018deep, yu2022surprising} and Unity ML-Agents Toolkit \cite{juliani2018unity}, such as batch size, learning rate, and epochs. These hyperparameters are listed in Table \ref{ppo-param}.} The {\color{black} versions of used training tools} are ml-agents-toolkit: 0.27.0, ml-agents-envs: 0.27.0, communicator API: 1.5.0, and PyTorch: 1.8.2+cu11.1. To accelerate the training process, $6$ copies of the environment are placed in one scene, and $3$ concurrent Unity instances (in total 18 parallel environments) are invoked at the start of the training. 

\begin{figure}[!tbp]
\setlength{\abovecaptionskip}{0.cm}
\setlength{\belowcaptionskip}{-0.cm}
\centering
\subfigure[]{
\includegraphics[width=1.62in,trim=10 0 40 20,clip]{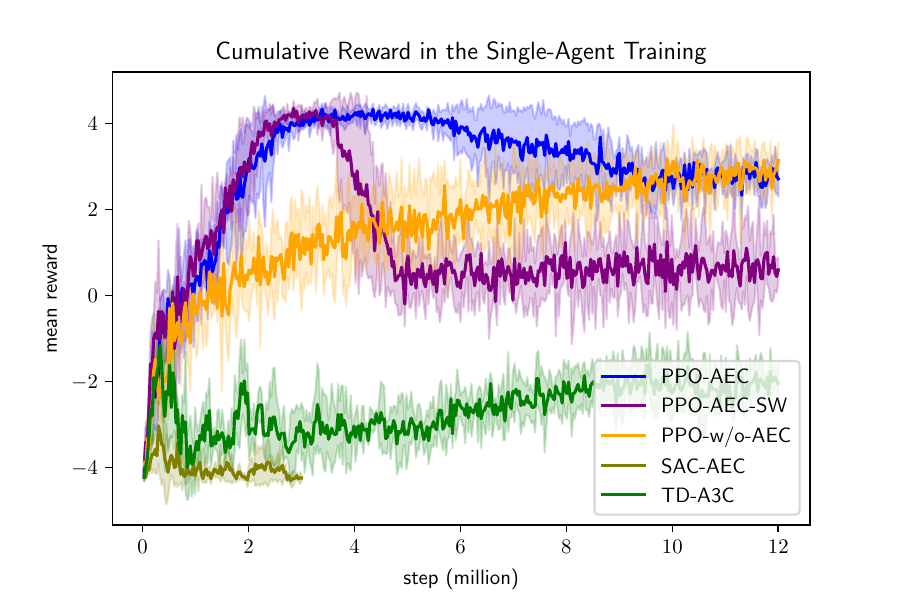}
}
\subfigure[]{
\includegraphics[width=1.62in,trim=10 0 40 20,clip]{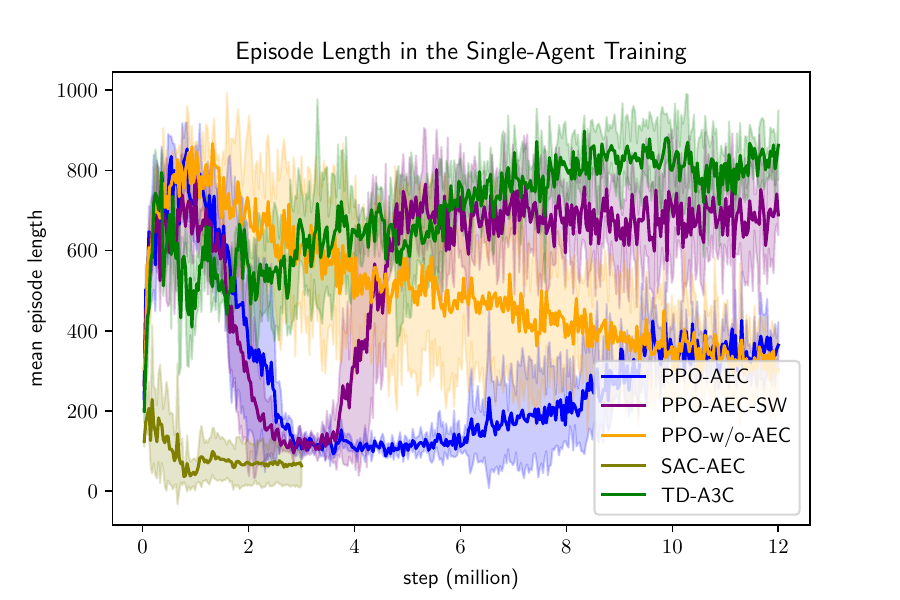}
}
\quad
\subfigure[]{
\includegraphics[width=1.62in,trim=10 0 40 20,clip]{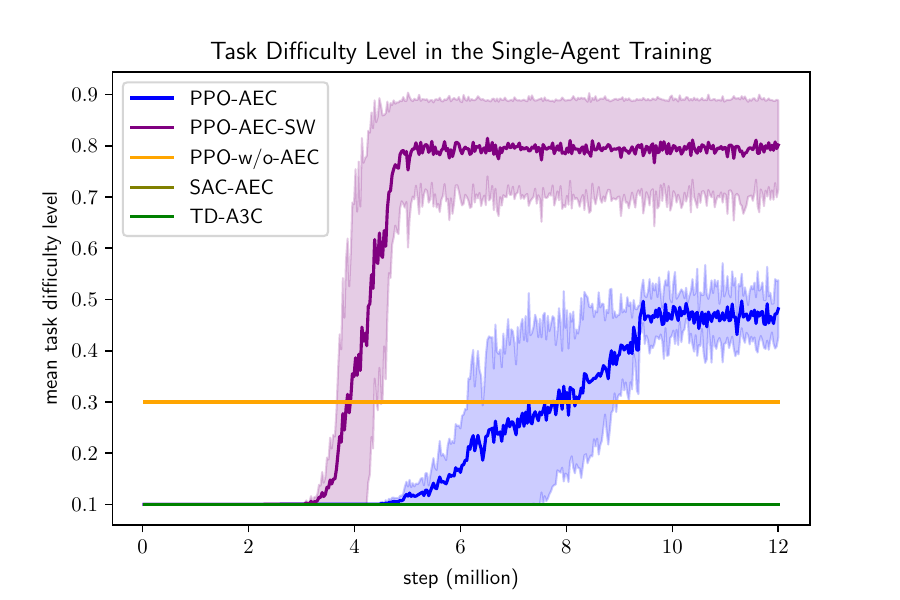}
}
\subfigure[]{
\includegraphics[width=1.62in,trim=10 0 40 20,clip]{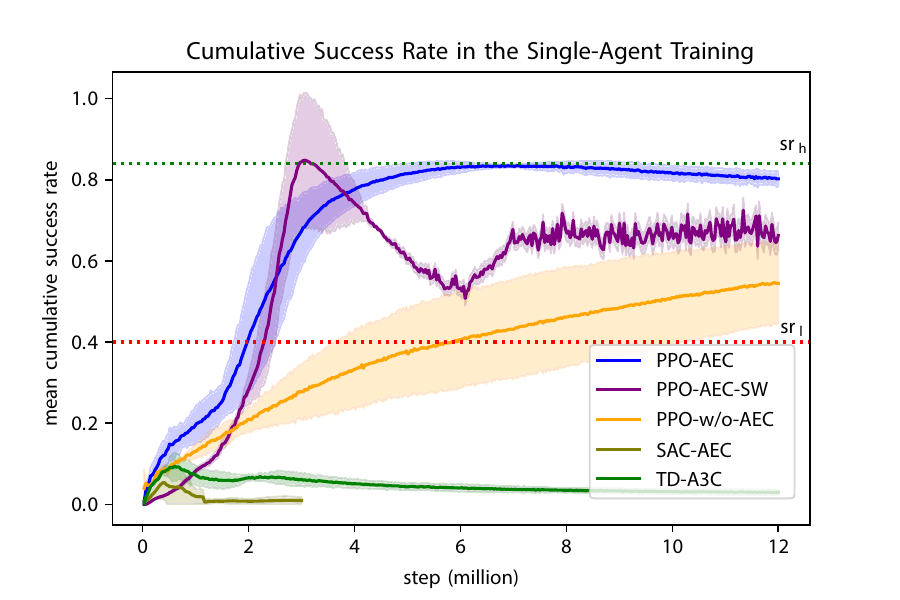}
}

      \caption{Learning curves of the PPO-AEC (ours), PPO-AEC-SW (ours), the baseline TD-A3C and the ablation PPO-w/o-AEC, SAC-AEC in the single-agent training. Mean and standard deviation (shaded) over $3$ independent runs conducted every 12 million training steps (episodes). {\color{black}Note that due to the high computational cost of SAC-AEC, we only consider 3 million training episodes for it. The moving SR is illustrated for PPO-AEC-SW in (d). All these values are the means over $3\times6 =18$ training environments. From (c) and (d), for PPO-AEC, when the cumulative {\color{black}SR} crosses the $sr_h=0.85$ ($4.5$ million steps), the TDL increases with $\delta_{\epsilon} = 0.1$ and converges to $0.48$ ($9.4$ million steps) with the SR of $80\%$.} (a) Cumulative reward; (b) Episode length; (c) Task difficulty level; (d) Cumulative success rate.}
      \label{single-train}
\end{figure}

The mean and the standard deviation values over $3$ runs of the single-agent training are illustrated in Fig. \ref{single-train}. As shown in Fig. \ref{single-train}, our PPO-AEC achieves the best performance even at a higher {\color{black}TDL ($\epsilon = 0.48$) compared to PPO-w/o-AEC and TD-A3C based on the {\color{black}convergent cumulative SR (80\%)} and the episode length {\color{black}(350)}. PPO-AEC achieves higher cumulative rewards (3.5) compared to PPO-w/o-AEC with the same TDL ($\epsilon$ =0.3), while also achieving the same rewards with a higher TDL. Our PPO-AEC-SW converges faster to a higher TDL ($\epsilon =0.8$) with {\color{black}the help of a sliding window} but induces a larger oscillation during training.} It verifies the data efficiency and effectiveness of the proposed AEC. {\color{black}TD-A3C and SAC-AEC cannot accomplish the search task as the {\color{black}SR} is below $10\%$ even with the lowest TDL ($\epsilon =0.1$). SAC-AEC obtains the worst performance as the policy diverges without improvement. Agent trained with TD-A3C can hardly find the target, illustrating the drawbacks of targeted-driven navigation with purely visual perception for target search.}

\begin{figure}[tbp]
\centering
\subfigure[]{
\includegraphics[width=1.62in,trim=10 0 40 20,clip]{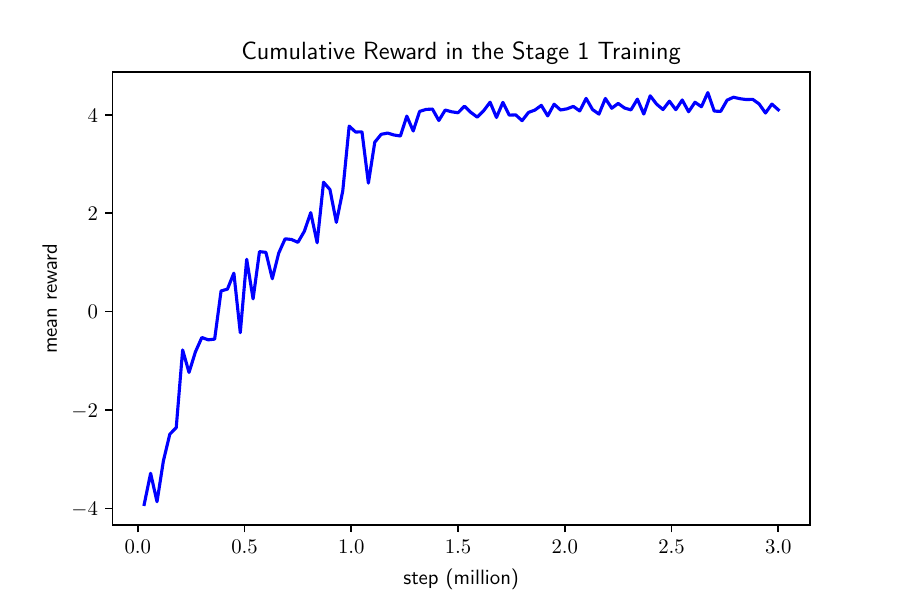}
}
\subfigure[]{
\includegraphics[width=1.62in,trim=10 0 40 20,clip]{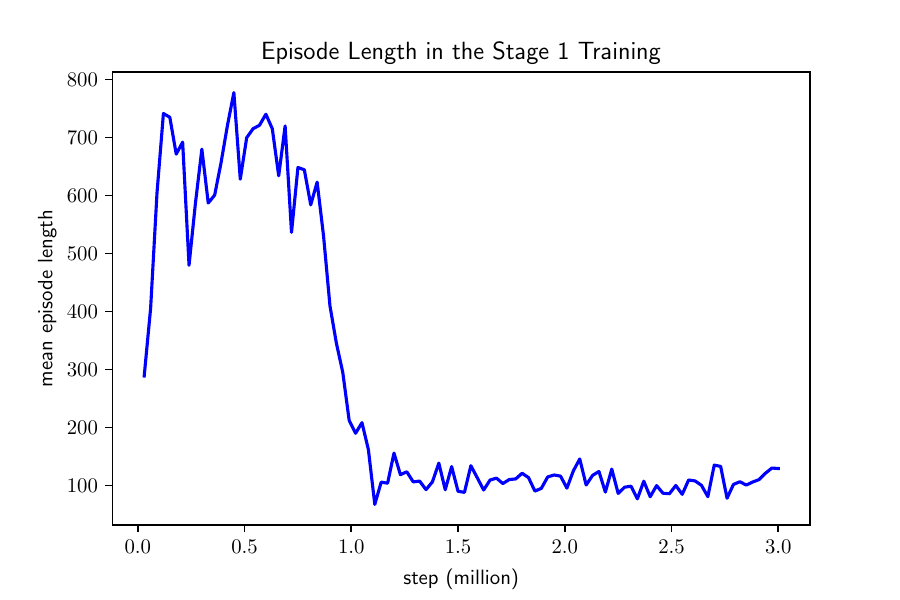}
}
\quad
\subfigure[]{
\includegraphics[width=1.62in,trim=10 0 40 20,clip]{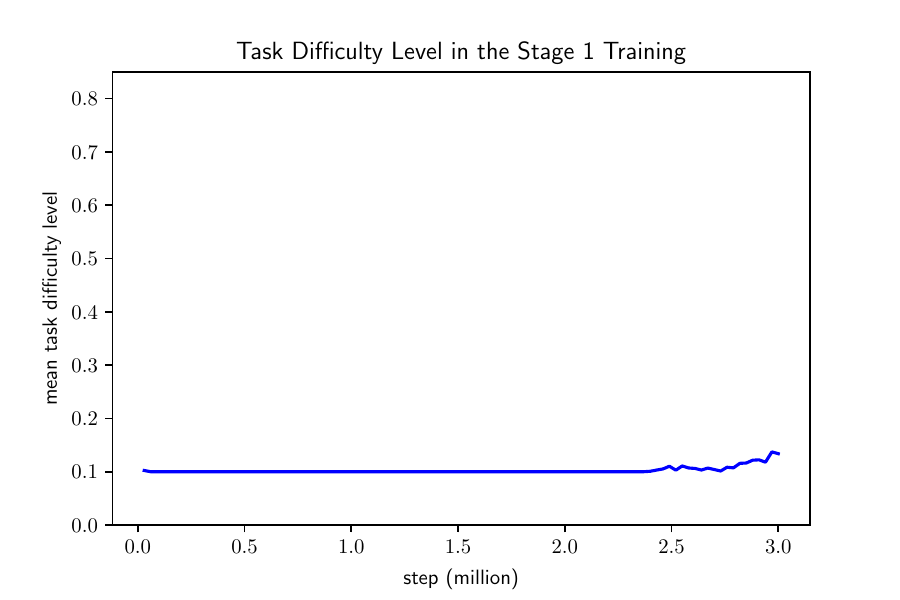}
}
\subfigure[]{
\includegraphics[width=1.62in,trim=10 0 40 20,clip]{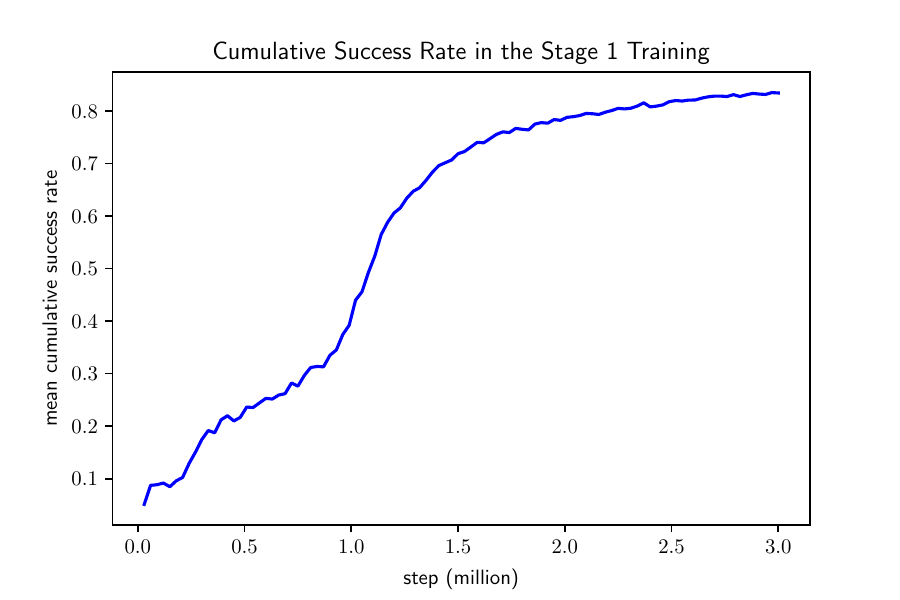}
}
\caption{Stage 1 training curves with single-agent over 3 million steps (episodes). {\color{black}Note that the same initial policy is used for both ACEMSL and CM3 in the Stage 2 training}. (a) Cumulative reward; (b) Episode length; (c) Task difficulty level; (d) Cumulative success rate.}
\label{stage1-train}
\end{figure}

\begin{figure}[tbp]
      \centering
\subfigure[]{
\includegraphics[width=1.62in,trim=0 0 0 0,clip]{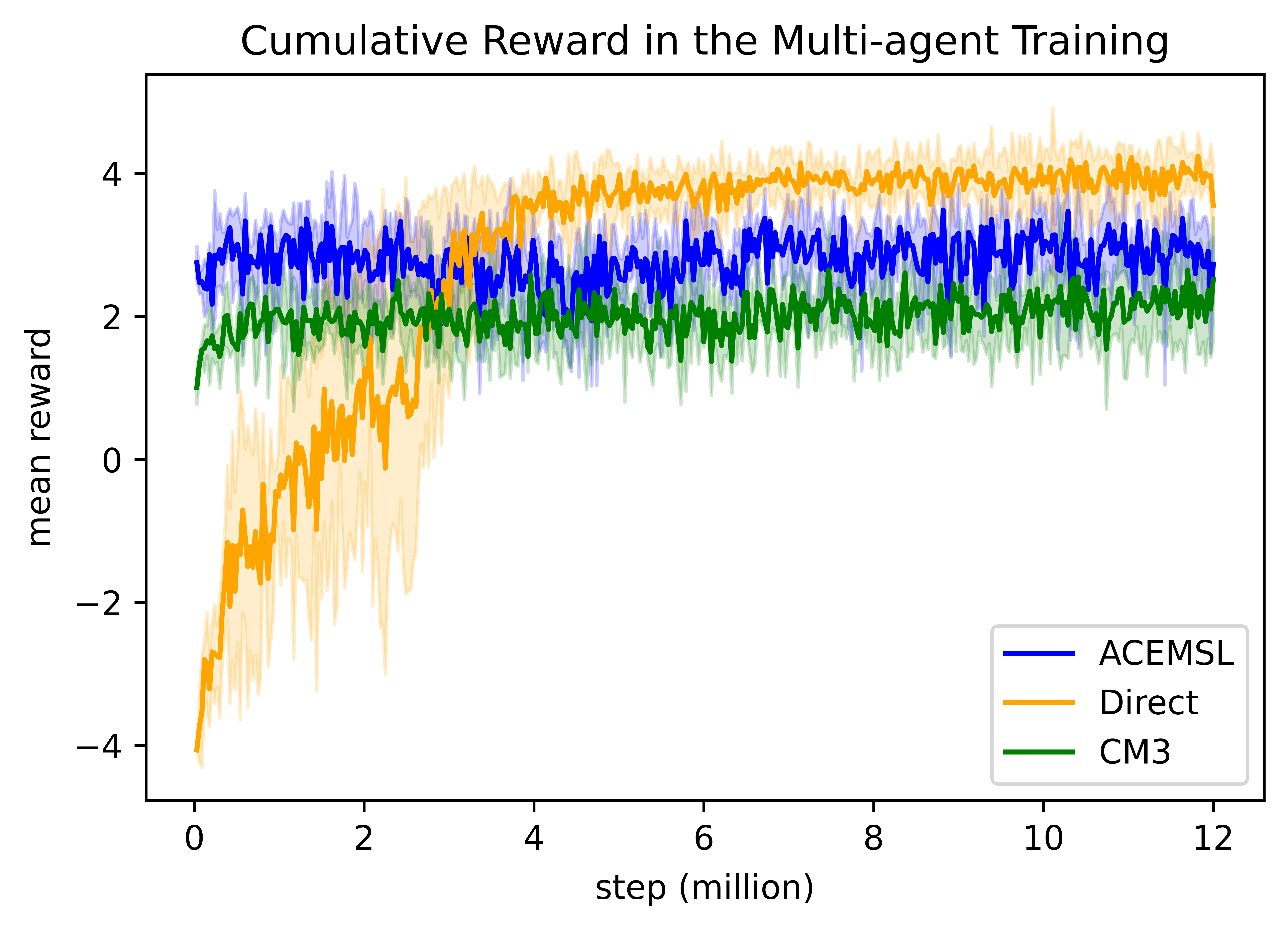}
}
\subfigure[]{
\includegraphics[width=1.62in,trim=0 0 0 0,clip]{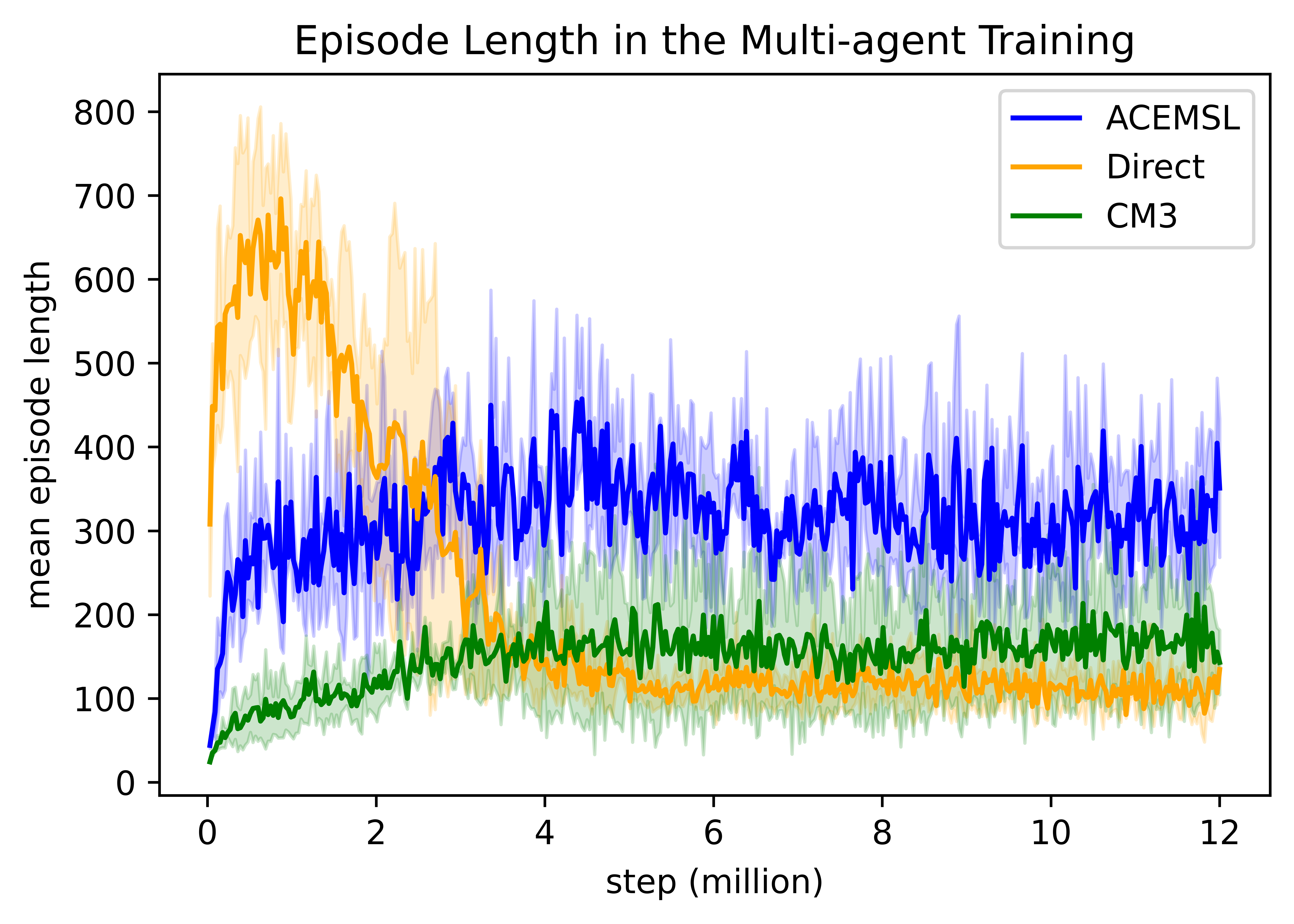}
}
\quad
\subfigure[]{
\includegraphics[width=1.62in,trim=0 0 0 0,clip]{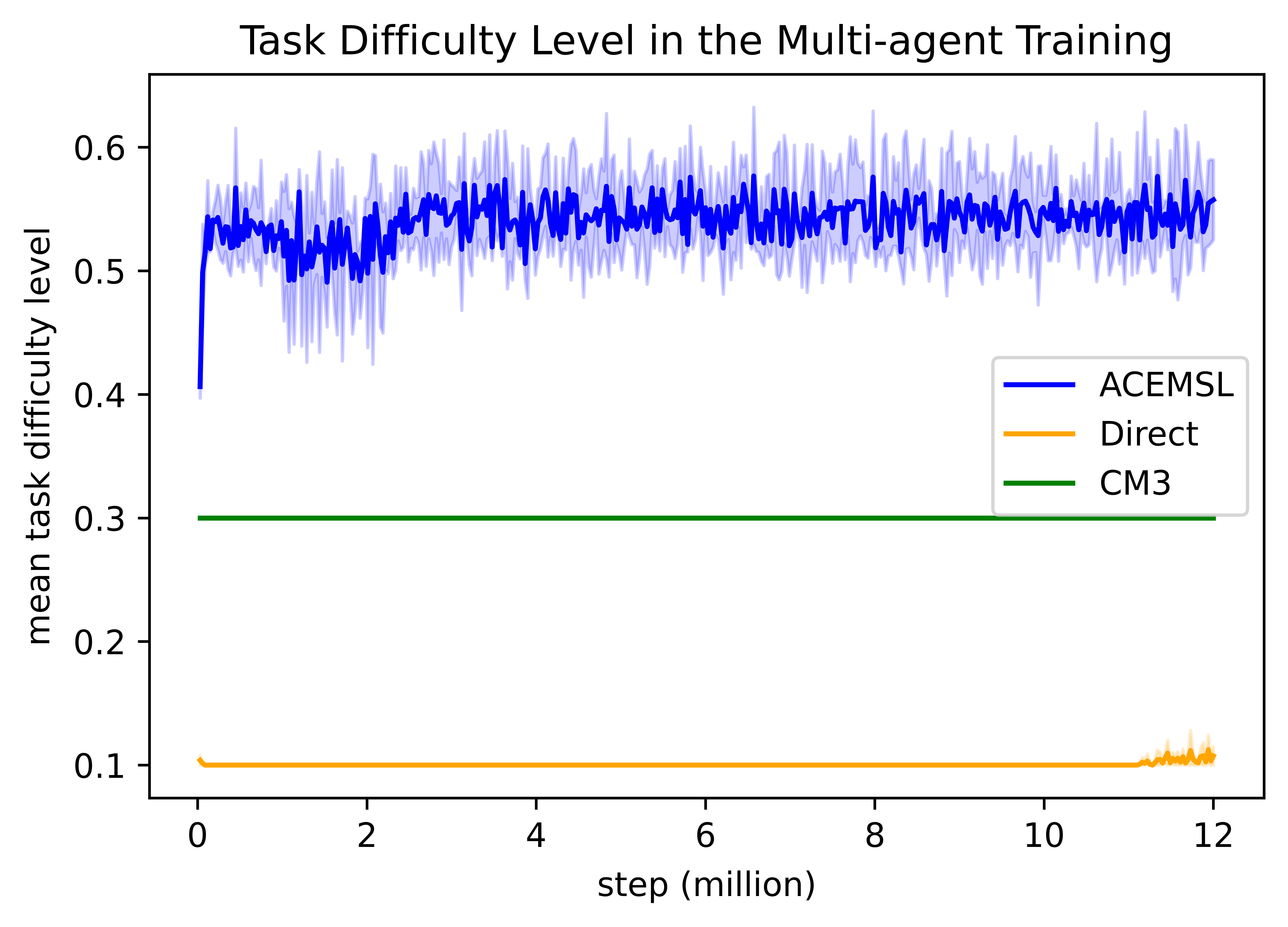}
}
\subfigure[]{
\includegraphics[width=1.62in,trim=0 0 0 0,clip]{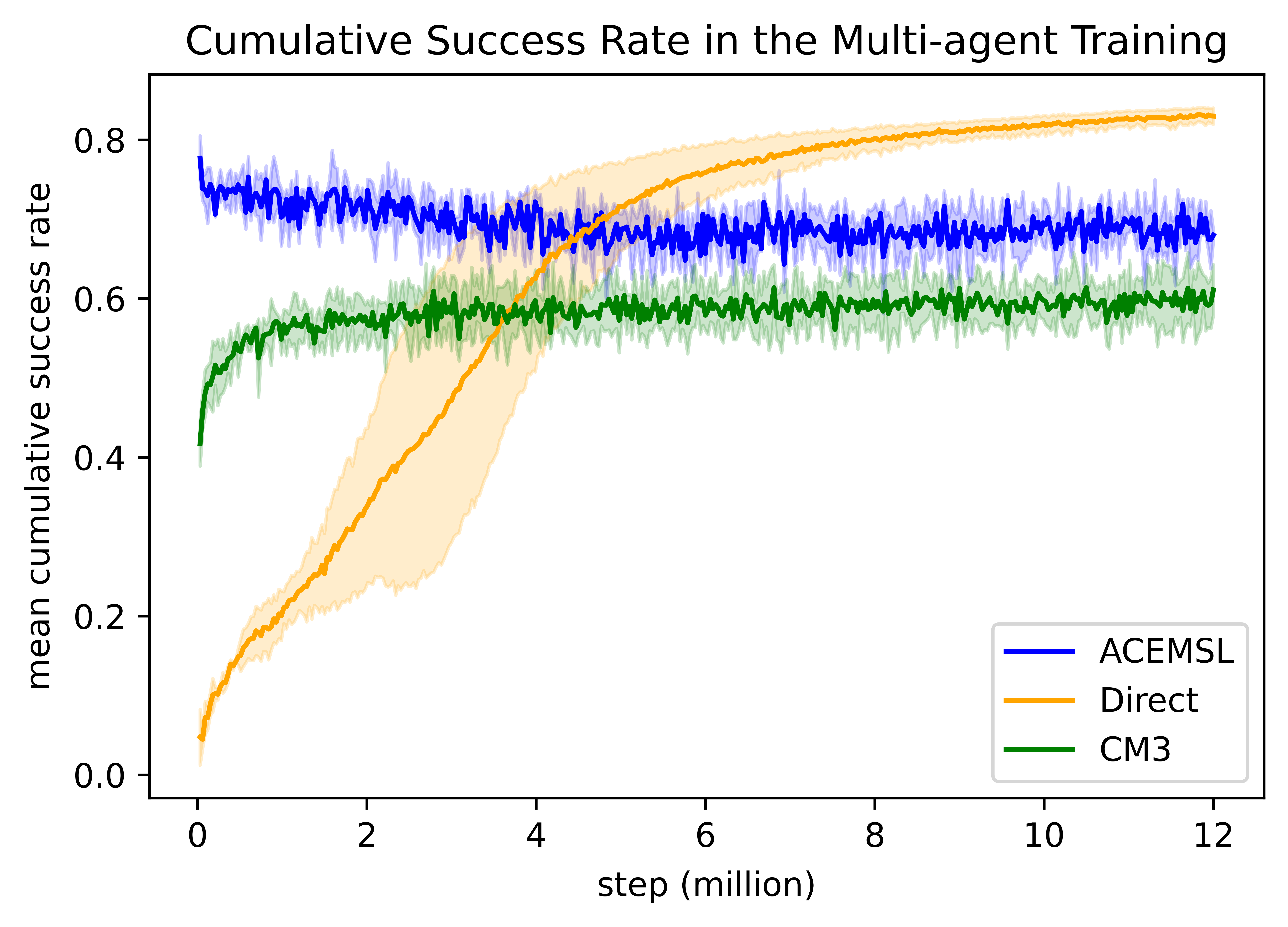}
}
      \caption{Learning curves of the ACEMSL (ours), the baseline CM3 and the ablation ``Direct" in the multi-agent training. Mean and standard deviation (shaded) over $3$ independent runs conducted every 12 million training steps (episodes). (a) Cumulative reward; (b) Episode length; (c) Task difficulty level; (d) Cumulative success rate.}
      \label{multi-train}
\end{figure}

For the multi-agent training, Stage 2 training is initialized from the existing model of Stage 1 trained over $3$ million {\color{black}episodes with PPO-AEC}. Stage 1 training curves are shown in Fig. \ref{stage1-train}. The mean values and the standard deviation over 3 runs of the multi-agent training are illustrated in Fig. \ref{multi-train}. With the help of Stage 1, ACEMSL and CM3 converge more than {\color{black} \textbf{1 million episodes} (total 3.1 million episodes: 3 million in Stage 1 and 0.1 million in Stage 2)} faster than the ``Direct" method {\color{black}(total 4.1 million episodes) as indicated in Fig. \ref{multi-train} (a)}, which demonstrates the significant {\color{black}data efficiency} of the multistage learning. Compared to CM3, ACEMSL benefits significantly from the AEC algorithm, as ACEMSL achieves a higher cumulative reward and {\color{black}SR}, even with higher TDLs (see Fig. \ref{multi-train} (c)). The AEC can fully liberate the advantages of multistage learning for CTS tasks with a visual drone swarm.

{\color{black}The proposed ACEMSL effectively addresses the challenges of 3D sparse continuous space exploration and the requirement for collaborative behavior in multi-agent systems, as demonstrated in both single-agent and multi-agent training. By combining the advantages of adaptive curriculum learning and multistage learning, our method significantly improves data efficiency in the training process.} {\color{black}Table \ref{runtime} lists the absolute training runtime for each of the algorithms. Note that Stage 1 training time is included in the CM3 and ACEMSL results. Our PPO-AEC algorithm for single-agent training and ACEMSL approach for multi-agent training cost average $(5807 \pm 80)~ s/(million ~episode)$ and $(5732 \pm 87)~ s/(million~ episode)$, respectively. {\color{black}From the obtained results, our ACEMSL framework outperforms CM3 in terms of the average training runtime metric.} Compared to the on-policy PPO algorithm, the off-policy SAC algorithm takes {\color{black}an average} $(305261 \pm 23614) ~s/ (million ~episode)$ during training due to the memory replay mechanism, which is not suitable for our applications. The converge training runtime of ACEMSL can be easily obtained from $(5732 \pm 87) ~s/(million~ episode) \times 4.1~ million~episodes = (23501.2 \pm 356.7)~ s \approx 6.53 ~hours$. From the analysis, we can conclude that ACEMSL can be applied to a large-scale drone swarm without much computational time cost.}

\begin{table}[tbp]
{\color{black}
\caption{Absolute Training Runtime of Different Algorithms (seconds)}
\label{runtime}
\centering
\resizebox{\columnwidth}{!}{
\begin{tabular}{@{}lccc@{}}
\toprule
Algorithms                & Episodes  &Absolute Runtime     & Average Runtime     \\ 
                          & (million) &(s)                  & (s/million episode) \\ \midrule
TD-A3C\cite{zhu2017target}  & 12 & 61023 $\pm$ 520 & \textbf{5085 $\pm$ 43}  \\
PPO-w/o-AEC \cite{schulman2017proximal}     & 12 & 66573 $\pm$ 1727 & 5548 $\pm$ 144 \\
SAC-AEC  & 3 & 915783 $\pm$ 70841& 305261 $\pm$ 23614 \\ 
\textbf{PPO-AEC (ours)} &12    &69689 $\pm$ 958    & 5807 $\pm$ 80\\ 
\textbf{PPO-AEC-SW (ours)} &12    &67186 $\pm$ 361    & 5599 $\pm$ 30\\
\midrule
CM3\cite{Yang2018}  & 15   & 100724 $\pm$ 2580 & 6715 $\pm$ 172 \\
Direct & 12  & 71893 $\pm$ 1320  & 5991 $\pm$ 110\\
\textbf{ACEMSL (ours)} & 15 & 85985 $\pm$ 1298 & \textbf{5732 $\pm$ 87}\\
\bottomrule
\end{tabular}
}
}
\end{table}

{\color{black}
\subsubsection{Validation}
To further validate the better performance of the proposed ACEMSL approach, we conducted inference experiments with trained policies in scenarios with different TDLs, namely $\epsilon = 0.3, \epsilon = 0.5, \epsilon = 0.7, \epsilon = 0.9$, at the predetermined initial position. The room dimension was the same as the training environment, i.e., $5\times5\times3$ $(m^3)$. The SR and the average TTR were calculated over 500 episodes for each approach. The results (see Table \ref{performance}) show that our proposed approaches achieve the best performance over all test TDLs in the single-agent and two-agent scenarios, respectively, in terms of SR. The use of the AEC-SW algorithm leads to improved performance compared to the AEC. Compared to PPO-AEC-SW, ACEMSL demonstrates that collaborative search significantly improves over single-agent search in terms of SR and TTR. While TD-A3C allows agents to avoid obstacles with a longer search time, it fails to find the target. SAC-AEC, on the other hand, fails to learn even an obstacle avoidance policy and only randomly touches the target with the least TTR as it diverges during training, as discussed in Section \ref{ppo-sac}. The pre-trained policy in Stage 1 also leads to better performance in Stage 2, as seen with the CM3 method outperforming the ``Direct" method. 

\begin{table}[tbp]
\begin{center}
{\color{black}
\caption{Performance (SR: \% / TTR: Avg (steps)) with Various Task Difficulty Levels (TDLs) in Simulation}
\label{performance}
\centering
\resizebox{\columnwidth}{!}{
\begin{tabular}{@{}lcccc@{}}
\toprule
Algorithms / TDL ($\epsilon$)    & 0.3  & 0.5     & 0.7 &  0.9     \\ \midrule
TD-A3C\cite{zhu2017target} & 5.4 / 2052 & 3.8 / 1434 & 0.6 / 1800 &  0.0 / N.A.\\
PPO-w/o-AEC\cite{schulman2017proximal}   & 57.1 / 129 & 43.9 / 142 & 39.3 / 468 & 25.1 / 829 \\
SAC-AEC & 0.4 / 255 & 0.4 / 77 &0.4 / 154  & 0.0 / N.A. \\ 
\textbf{PPO-AEC (ours)} &70.5 / 180    & 52.9 / 240    & \textbf{48.7} / 295 & 31.7 / 322\\ 
\textbf{PPO-AEC-SW (ours)} &\textbf{74.7} / 158    & \textbf{63.7} / 192    & 47.1 / 208 & \textbf{32.7} / 312\\ \midrule
CM3\cite{Yang2018}  & 74.7 / 154   & 64.3 / 216 & 50.1 / 232 &  37.3 / 384 \\
Direct & 69.1 / 244  & 52.1 / 333 & 35.3 / 357 & 22.6 / 522 \\
\textbf{ACEMSL (ours)} & \textbf{79.3 / 107} & \textbf{66.5 / 150} & \textbf{51.5 / 182} & \textbf{38.5 / 297}\\
\bottomrule
\end{tabular}
}
}
\end{center}
\end{table}

\begin{figure*}[!tbp]
      \centering
      \includegraphics[width=7in]{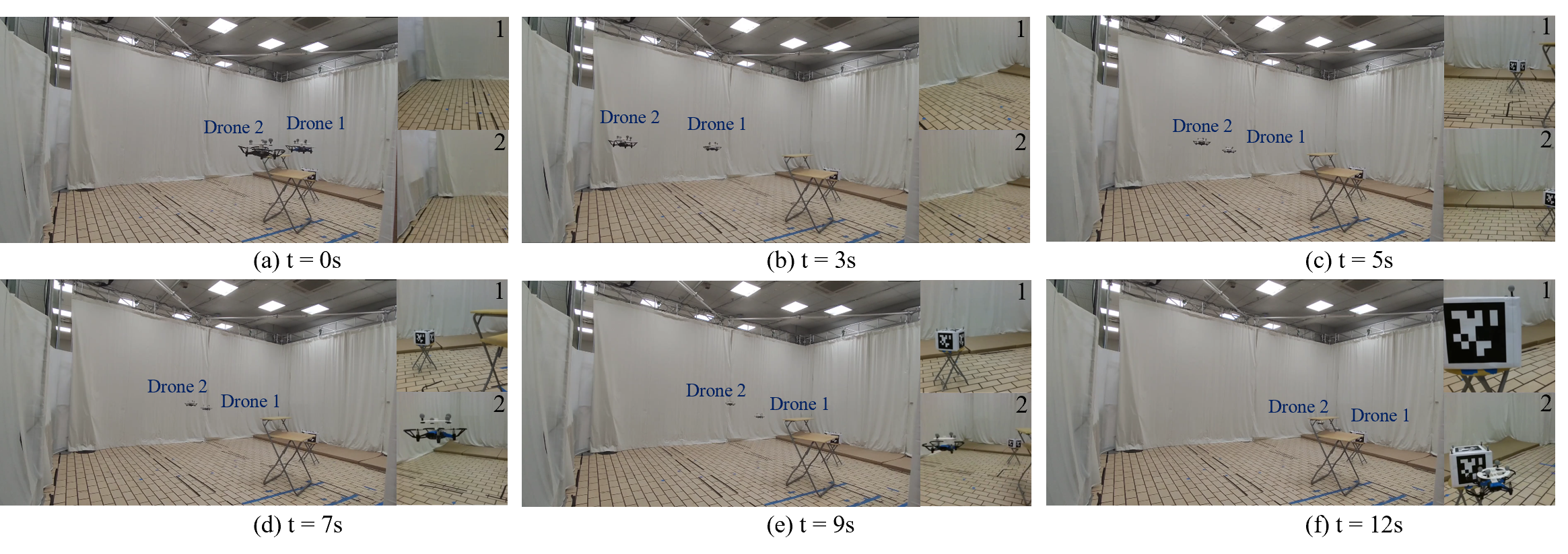}
      \caption{The snapshots of the visual drone swarm {\color{black}performing} the CTS. The images perceived by the cameras of Drone 1 and Drone 2 are labeled as 1 and 2, respectively at the top right side of each snapshot.}
      \label{snapshot}
\end{figure*}

\subsubsection{Testing}
We tested our models from single-agent and multi-agent training on various room dimensions with $500$ episodes evaluated per experiment, where the {\color{black}TDL} is set as $\epsilon = 0.3$. We also demonstrated the scalability of the trained model with ACEMSL, where multiple drones are spawned {\color{black}in} the start area. The {\color{black}SR} and the mean {\color{black}TTR with different number} of drones on various room dimensions are listed in Table \ref{tests}. As the room dimensions increase, the {\color{black}SR} drops, and the required time steps to reach the target increase. On the other hand, deploying more drones helps increase the {\color{black}SR} and reduce the required time steps to reach the target, which illustrates the advantages of using a visual drone swarm for target search. However, the scale of the swarm is constrained by the size of the room (the {\color{black}SR} drops from 75.4\% to 63.8\% in Table \ref{tests} when 3 drones are deployed in the room of $5m\times 5m \times 3m$) since there is a higher chance of the drones crashing into one another if too many drones are flying within a small room. 

\begin{table}[tb]
\begin{center}
\caption{Performance of ACEMSL ({\color{black}SR:} \% / {\color{black}TTR: Avg (steps)}) with Various Room Dimensions {\color{black}in Simulation}}
\label{tests}
\centering
\begin{tabular}{@{}lccc@{}}
\toprule
Room Dimensions ($m$)               &  $5\times5\times3$ & $8\times8\times3$    & $10\times10\times3$         \\ \midrule
1  drone   &  75.4\% / 162           &    52.8\% / 301          &    16.0\% / 325            \\
2  drones & \textbf{79.3\%} / 107 & 61.1\% / 236 & 35.9\% / 256   \\
3  drones      & 63.8\% / \textbf{104} & \textbf{62.0\% / 178} & \textbf{41.9 \% / 254}  \\
\bottomrule

\end{tabular}
\end{center}
\end{table}

\subsection{Physical Experiments}
To evaluate the trained models' generalization and Sim2Real {\color{black}capabilities}, we transfer our models to an unseen large indoor environment without fine-tuning, and their performance on different devices is compared. The test environment is an enclosed room with the size of $8m \times 7m \times 4m$. Each side, {\color{black}excluding the floor and ceiling of the environment, is covered with white curtains. The target is a parcel box with the size of $0.2m \times 0.3m \times 0.2m$, placed randomly in/behind obstacles within two environments in several test positions such as edges and corners to control the TDL near $\epsilon=0.5$ (Env 1, fewer obstacles) or $\epsilon=1.0$ (Env 2, all choosing from the set $U(\bm{C})$ with more obstacles; for detailed locations, see the project website).} Each side of the box is attached with a printed AprilTag (Tag36h11) \cite{olson2011tags}. An OptiTrack motion capture system with eight cameras is used {\color{black}to broadcast} the positions of the drone(s) and the target. The position of the target is used to detect whether the drones can achieve their goal. In our experiment, the laptop computing center and Tello Edu drones are used to deploy our trained models and conduct flight tests. In addition, for deployment on the onboard computers, three of the following computing platforms are evaluated on their performance in terms of the execution speed of the trained model using only their CPUs. Note that the trained model for only a single drone deployment is considered for this evaluation.
\begin{itemize}
\item Lenovo Ideapad Slim 5 Pro laptop running Ubuntu OS with AMD Ryzen 7 5800H CPU and 16GB DDR4 RAM.

\item NVIDIA Jetson Xavier NX running Ubuntu OS with {\color{black}a} 6-core NVIDIA Carmel 64-bit CPU and 8GB RAM.

\item Raspberry Pi 4 Model B with 64-bit ARM Cortex-A72 (ARMv8) and 8GB RAM.
\end{itemize}

All of these devices use Python3 and ONNXRuntime\footnote{https://onnxruntime.ai/} for inference with the trained model in ONNX format. PyTorch and Torchvision are used for data processing on all devices. Table \ref{devices} illustrates the execution speed of the trained model in units of frames per second (fps).

\begin{figure}[tbp]
      \centering
      \includegraphics[width=3.3in, trim=0in 0.05in 0in 0in, clip]{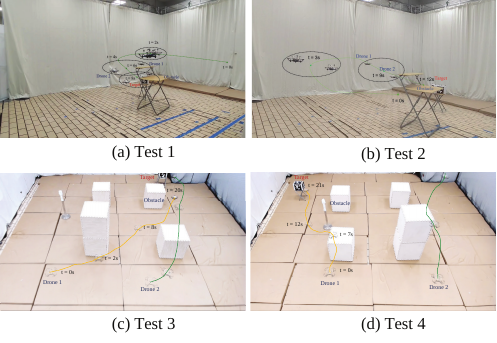}
      \caption{The trajectories of the drone swarm after successfully performing the CTS in four tests with different initial conditions {\color{black}(Test 1-2 from Env 1 and Test 3-4 from Env 2).} The stability of the search policy is demonstrated with different initial positions and forward directions for the drones and the target.}
      \label{traj}
\end{figure}

\begin{table}[tbp]
\centering
\caption{Computing Performance (Avg $\pm$ Std) of Devices}
\label{devices}
\begin{tabular}{@{}lccc@{}}
\toprule
Devices                & Raspberry Pi & Xavier NX    & Laptop         \\ \midrule
Program Component     &              &              &                \\
Image Processing (ms)  & 85.1 $\pm$ 3.8 & 35.9 $\pm$ 0.6 & 16.1 $\pm$ 0.6   \\
ONNX Model (ms)        & 48.2 $\pm$ 2.8 & 10.5 $\pm$ 0.4 & 4.47 $\pm$ 0.49  \\
Control Commands (ms)  & 3.08 $\pm$ 0.75& 1.62 $\pm$ 0.15& 0.35 $\pm$ 0.03  \\ \midrule
All Process \\
(ms)                   & 128 $\pm$ 7 & 48.1 $\pm$ 0.9 & \textbf{19.8 $\pm$ 0.6}   \\
(fps)                  & 8.15 $\pm$ 0.33& 20.9 $\pm$ 0.4 & \textbf{51.2 $\pm$ 1.3}  \\ 
\bottomrule

\end{tabular}
\end{table}

Real-time processing capability is crucial for model deployment on computing devices. The evaluation criterion for device performance is set such that for any device {\color{black}with} real-time processing capability for neural network models, it should be able to execute every part of the trained models, including the data pre-processing step, with {\color{black}a} speed of at least 20 frames per second (fps). According to the {\color{black}results} in Table \ref{devices}, only the NVIDIA Jetson Xavier NX and Lenovo Ideapad Slim 5 laptop meet the evaluation criterion; therefore, these devices have real-time processing capabilities. However, the Raspberry Pi execution speed is too low to be considered real-time processing. As the experiment evaluated only the performance using CPU, the performance of NVIDIA Jetson Xavier NX and Lenovo Ideapad Slim 5 laptop can be improved further using GPUs, parallel computing with CUDA, and deep learning accelerators, such as TensorRT\footnote{https://developer.nvidia.com/tensorrt}.

Each drone is {\color{black}configured} as a wireless AP connected to the computing center during real-time flight tests. As every Tello Edu drone has a similar, immutable IP address (192.168.10.1) on its network, there is {\color{black}an issue of IP address conflict} when all drones are connected {\color{black}to} the same computer. To resolve, network routing is required to route drones' network to different virtual IP addresses perceived by the computer via docker services\footnote{https://www.docker.com/}.
\begin{figure}[tbp]
      \centering
      \includegraphics[width=3.3in]{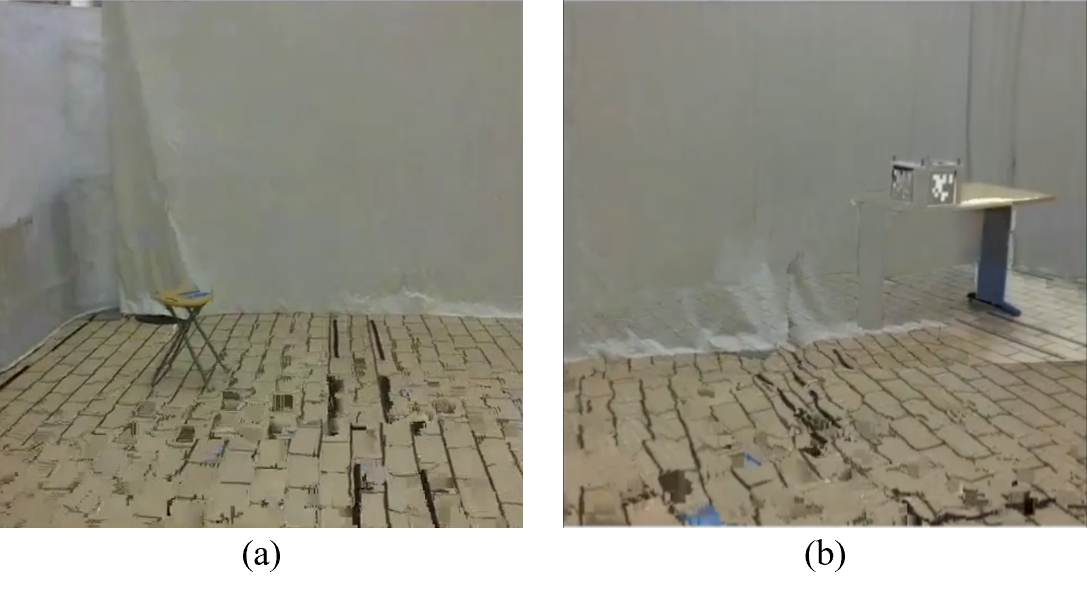}
      \caption{Image {\color{black}blur} and loss of the video stream cause the failure of tasks. (a) Obstacle {\color{black}blur}; (b) Target {\color{black}blur}.}
      \label{blur}
\end{figure}

\begin{table}[!tb]
\begin{center}
\caption{Target Search Performance in {\color{black}Physical Experiments}}
\label{accuracy}
{\color{black}
\begin{tabular}{@{}lcc|cc@{}}
\toprule
TDL       & \multicolumn{2}{c|}{Env 1, $\epsilon=0.5$} & \multicolumn{2}{c}{Env 2, $\epsilon=1.0$}\\ 
Agents & Single Drone & Two Drones  &  Single Drone & Two Drones               \\ \hline
{\color{black}SR}       & 48\%           & \textbf{64\%}    & 28\% &    \textbf{44\%}   \\ 
\hline
{\color{black}TTR}     &                &                                                                                              \\
(second)           & 12.9 $\pm$ 0.8 & \textbf{12.2 $\pm$ 0.8}    & $45.4\pm 14.9$  &  \textbf{39.1 $\pm$ 9.2}                                                                        \\
(step)             & 143 $\pm$ 10   & \textbf{142 $\pm$ 13}  & $656 \pm 236$  & \textbf{567 $\pm$ 124} \\
\bottomrule
\end{tabular}
}
\end{center}
\end{table}

The drones stream the video data captured by their cameras to the laptop computing center via WIFI. PyAV library is imported to open and decode drones' video stream into image frames with a size of $3\times720\times960$ recognized by Torchvision. Considering the mismatch between the aspect ratios of the raw images and the model training input, the image frames are firstly cropped to $3\times720\times720$ around the image center. Thenceforth, Torchvision reshapes and converts {\color{black}each cropped image} into a NumPy array of $3\times224\times224$, consistent with the size of visual observation in the trained model. {\color{black}The model generates translational and angular velocity commands in a loop block based on the obtained visual data and the drone's position.} These velocity commands drive connected drones to search for and reach the target collaboratively in real-time (see Fig. \ref{snapshot}).

In this work, we conducted 25 tests each for {\color{black}single-drone} and collaborative {\color{black}two-drone} target search to evaluate the performance of trained models in an unseen physical environment. This evaluation focuses on SR and TTR. The experiments for single-drone and collaborative two-drone target search use different models from the single-agent and multi-agent training, respectively. During the tests, the drones were randomly placed in the start area with varying forward directions and positions. {\color{black}A test is considered a success if a drone reaches and lands within $0.5~m$ of the target; otherwise it is considered a failure.} Fig. \ref{traj} shows the trajectories of several successful test cases.

{\color{black}Table \ref{accuracy} compares the performance of the target search using a single drone and two collaborative drones in different physical experiments with $\epsilon = 0.5$ and $\epsilon = 1.0$. The results show that the CTS has higher performance in terms of TTR and SR ($64\%/44\%$ for two drones, compared to $48\%/28\%$ for a single drone).} This indicates that collaboration among drones helps to find the target faster and more accurately. {\color{black}Furthermore, the SR in physical experiments is close to that in simulations though in unseen environments,} as shown in Table \ref{tests}, demonstrating the generalization and Sim2Real capabilities of trained models. More details can be found in \url{https://github.com/NTU-UAVG/CTS-visual-drone-swarm.git}. 

{\color{black}
\subsection{Analysis and Discussion}
\subsubsection{Stability Analysis}
{\color{black}Although} drones with the trained optimal policy receive feedback from visual perception, the explicit relative position between the target and the agents cannot be obtained in our scenarios. Since the cascade controller is adopted and we only focus on the high-level search decision, the system can be formulated as
\begin{equation}
    \dot{\bm{x}} = \bm{g(x)} \cdot \bm{u}
\end{equation}
where $\bm{x} = [x,y,z,\psi]^T$ with $\psi$ as yaw angle, $\bm{g(x)} = [\cos{\psi}, \sin{\psi}, 0, 0; \sin{\psi}, \cos{\psi}, 0,0;0,0,1,0;0,0,0,1]$, and the output command of trained policy $\bm{u} =[\hat v_x^B, \hat v_y^B, \hat v_z^B, \hat \omega_z]^T$. {\color{black}Assuming} that the image encoder can extract the relative position $\bm{x_d}$ once the target is recognized, we choose a Lyapunov function $\mathcal{L} = \frac{1}{2}\bm{e}^T\bm{e}$ with $\bm{e} = \bm{x}-\bm{x}_d$. The time derivative $\dot{\mathcal{L}} = \bm{e}^T\bm{g(x)}\bm{u} \leq 0$ if $\bm{u} = -\bm{K} \bm{g(x)}^T \bm{e}$. Hence, if the output of the trained policy has negative feedback of the state error $\bm{e}$ with the positive constant gain $\bm{K}$, the closed-loop system is stable. An in-depth analysis can be conducted with the help of the interpretability of the image encoder in our future work. During the inference process, the drones are able to continue the search task when their lost visual perception is recovered, as observed in our experiments. The stability capability of the search policy is also demonstrated in Fig. \ref{snapshot}, where drones start searching with initial state perturbations.

\subsubsection{Generalization Capability Analysis}
The proposed ACEMSL, which utilizes various domain randomization techniques, including environment randomization and agent randomization, is {\color{black}capable of transferring} trained policies from simulation to real drones performing CTS in a totally different physical environment. The generalization capability can be further improved by using more domain randomization techniques, such as wall texture and image noise filtering. Moreover, the use of relative position observation with the shortest distance allows for easy deployment of the trained policy in a large-scale drone swarm without additional training, unlike the function augmentation in CM3 \cite{Yang2018}.

\subsubsection{Failure Case Analysis}
In this work, we aim to achieve the best performance for CTS in physical environments using a visual drone swarm. The lack of global map memory, the position of the target, and the limited visual perception make our CTS task significantly more challenging compared to existing environment exploration problems \cite{8929168,9244647}, particularly in real-world environments. Despite these challenges, our approach is still able to achieve the best SR ($64\%$) in a $8m \times 7m \times 4m$ room. This is particularly impressive considering that the SOTA target-driven visual navigation approach \cite{9785445} only achieves an SR of $43\%$ along a $5\sim 10m$ path with a fixed room layout. 

Some of the failure cases can be attributed to the following factors: (1) video blur, as shown in Fig. \ref{blur}, which provides incorrect information to the trained model, causing the drones to fail to detect the target during flight; (2) imbalanced training dataset for target recognition in unseen corners and seen regions; (3) insufficient memory of the policy model without reserved map buffer. To address these issues, we can introduce the concept of adaptive loss function \cite{7859408} and anomaly detection \cite{9764628}, which {\color{black}improves} the robustness of the model to video with losses and noise. Additionally, similar to \cite{8698218}, we can utilize the spatio-temporal signal property of image sequences to retrieve target information from blurry images during runtime. Furthermore, we can apply the Recurrent Convolutional Attention (RCA) network \cite{8767027} to the imbalanced training dataset to improve environment understanding and memory during training and inference, which can help the drone infer the position and presence of the target in hidden corners. These techniques can improve the SR of our proposed approach for CTS tasks.
}

\section{Conclusion}
{\color{black}This paper proposes a data-efficient RL-based approach called ACEMSL to address the challenges of collaborative target search (CTS) for a visual drone swarm in a cluttered 3D space of sparse reward. Our approach improves upon existing curriculum learning by introducing a novel AEC algorithm and integrating it into task-specific multistage learning. Without prior knowledge of the target, the drone swarm performs CTS using only local visual perception and egocentric observations. Simulation results show that ACEMSL outperforms SOTA direct RL methods and the multistage learning CM3 in policy optimality, training runtime, as well as search efficiency. Moreover, we also demonstrate the generalization and Sim2Real capabilities of our trained models with real-world flight tests in an unseen physical environment. The experiment results verify the effectiveness of a visual drone swarm performing CTS based on the metrics of SR and TTR. To the best of our knowledge, this work is the first to achieve and demonstrate CTS with a visual drone swarm in the real world. 

The failure cases of our DRL framework motivate future works such as safe DRL with higher bandwidth to evaluate and improve the CTS performance under communication loss, image blur, and even cyber attacks on visual perception.}

\section*{Acknowledgment}

The authors would like to thank Chun Wayne Lee and Yi Hong Ng are with Nanyang Technological University, Singapore, for their help and valuable feedback on this work.

\ifCLASSOPTIONcaptionsoff
  \newpage
\fi



%
\bibliographystyle{IEEEtran}
\bibliography{IEEEabrv,ref}



%


\end{document}